\documentclass[twocolumn, switch]{article} 
\pdfoutput=1

\usepackage{preprint}

\usepackage{amsmath, amsthm, amssymb, amsfonts}

\usepackage[numbers,square]{natbib}
\usepackage[utf8]{inputenc}	
\usepackage[T1]{fontenc}	
\usepackage{xcolor}		
\usepackage[colorlinks = true,
            linkcolor = purple,
            urlcolor  = blue,
            citecolor = cyan,
            anchorcolor = black]{hyperref}	
\usepackage{booktabs} 		
\usepackage{nicefrac}		
\usepackage{microtype}		
\usepackage{lineno}		
\usepackage{float}			

\usepackage{lipsum}		
\usepackage{graphicx}
\usepackage{subfigure}
\usepackage{algorithm}
\usepackage{algorithmic}
\usepackage{makecell}

\usepackage{newfloat}
\DeclareFloatingEnvironment[name={Supplementary Figure}]{suppfigure}
\usepackage{sidecap}
\sidecaptionvpos{figure}{c}

\usepackage{titlesec}
\titlespacing\section{0pt}{12pt plus 3pt minus 3pt}{1pt plus 1pt minus 1pt}
\titlespacing\subsection{0pt}{10pt plus 3pt minus 3pt}{1pt plus 1pt minus 1pt}
\titlespacing\subsubsection{0pt}{8pt plus 3pt minus 3pt}{1pt plus 1pt minus 1pt}

\title{Towards Palmprint Verification On Smartphones}

\usepackage{xwatermark}
\newwatermark[firstpage,color=gray!90,angle=0,scale=0.28, xpos=0in,ypos=-5in]{*corresponding author: \texttt{cslinzhang@tongji.edu.cn}}

\usepackage{authblk}

\author[1,2]{Yingyi Zhang}
\author[1\thanks{\tt{cslinzhang@tongji.edu.cn}}]{Lin Zhang}
\author[2]{Ruixin Zhang}
\author[2]{Shaoxin Li}
\author[2]{Jilin Li}
\author[2]{Feiyue Huang}
\affil[1]{the School of Software Engineering, Tongji University}
\affil[2]{Youtu Lab, Tencent}

\begin{document}

\twocolumn[ 
  \begin{@twocolumnfalse} 

\maketitle

\begin{abstract}
  With the rapid development of mobile devices, smartphones have gradually become an indispensable part of people's lives. Meanwhile, biometric authentication has been corroborated to be an effective method for establishing a person's identity with high confidence. Hence, recently, biometric technologies for smartphones have also become increasingly sophisticated and popular. But it is noteworthy that the application potential of palmprints for smartphones is seriously underestimated. Studies in the past two decades have shown that palmprints have outstanding merits in uniqueness and permanence, and have high user acceptance. However, currently, studies specializing in palmprint verification for smartphones are still quite sporadic, especially when compared to face- or fingerprint-oriented ones. In this paper, aiming to fill the aforementioned research gap, we conducted a thorough study of palmprint verification on smartphones and our contributions are twofold. First, to facilitate the study of palmprint verification on smartphones, we established an annotated palmprint dataset named MPD, which was collected by multi-brand smartphones in two separate sessions with various backgrounds and illumination conditions. As the largest dataset in this field, MPD contains 16,000 palm images collected from 200 subjects. Second, we built a DCNN-based palmprint verification system named DeepMPV+ for smartphones. In DeepMPV+, two key steps, ROI extraction and ROI matching, are both formulated as learning problems and then solved naturally by modern DCNN models. The efficiency and efficacy of DeepMPV+ have been corroborated by extensive experiments. To make our results fully reproducible, the labeled dataset and the relevant source codes have been made publicly available at \url{https://cslinzhang.github.io/MobilePalmPrint/}.
  \end{abstract}
\vspace{0.35cm}

  \end{@twocolumnfalse} 
] 



\section{Introduction}\label{sect:Intro}

Smartphones have become a necessity in people's lives and an important tool to perform daily work. With the increasing concern about networking and mobility security, the demand for reliable user authentication technology is also growing dramatically \cite{jain2007handbook}. Propelled by the demand of various mobile applications, such as online payment and electronic banking, automatic identification of highly trusted people has become an intense study\cite{neal2016surveying}. In fact, high-precision identity authentication for smartphones has long been one of the main focuses in the research field, such as fingerprint-based unlocking \cite{huawei.com} and face-based e-payment \cite{apple.com}.

However, ignored by many, palmprint is also a great choice for personal identity authentication. Palmprint \cite{zhang2003online,zhang20153d,javidnia2015palm} refers to the skin pattern on the inner surface of the palm, which mainly includes two characteristics: palm frictional ridges (ridges and valleys like fingerprints) and palm flexion creases (discontinuities in epidermal ridges).
Palmprint, as an important member of the biometrics family, has many desired characteristics, such as strong uniqueness, durability and user-friendliness \cite{zhang2017towards}. The ability of palmprints to discriminate different identities has been confirmed by numerous previous studies \cite{meng2014surveying,zhong2019decade}.
In fact, palmprint authentication is a non-invasive way of authentication, which is more acceptable to users than facial authentication. Comparing fingerprint recognition with palmprint recognition, the former usually requires highly sensitive sensors, while the latter can be conducted by using the built-in camera of smartphones. In addition, some people do not have clear fingerprints, while palmprints have more abundant feature information and can be utilized in low-resolution conditions.

Although palmprint has many valuable merits as a biometric identifier, studies on palmprint recognition for smartphones are still quite rare, and consequently, the application potential of palmprints for smartphones is far from being fully explored. We attempt to fill this research gap by conducting a thorough study of palmprint verification on smartphones in this paper. A highly efficient and effective palmprint verification system specially designed for smartphone platforms is developed. In addition, a large-scale palmprint dataset collected by multi-brand smartphones is established and released to facilitate the further study of this area.

The remainder of this paper is organized as follows: Section \ref{sect:RelatedWorkContri} introduces related works and our contributions. Section \ref{sect:deepMPV} presents our DCNN-based palmprint verification system DeepMPV+ (short for ``Lighter Deep-learning-based Mobile Palmprint Verification''). Section \ref{sect:MPD} states the details of MPD (short for ``Mobile Palmprint Database''), which is our newly established benchmark dataset for the study of palmprint verification on smartphones.
Experimental results are reported in Section \ref{sect:Experiments}. Finally, Section \ref{sect:Conclusion} concludes the paper.

\section{Related Works and Our Contributions}\label{sect:RelatedWorkContri}

In this section, we will first review some representative studies most relevant to our work, including approaches for palmprint ROI extraction, approaches for palmprint verification, and publicly available benchmark datasets. Then, we will present our motivations and contributions.

\subsection{Approaches for Palmprint ROI Extraction}\label{subsect:roiExtraction}

Most pipelines for palmprint recognition need to segment hands from the background before ROI extraction.
Because of the unpredictable background and illumination of a mobile image, it is a challenging task to segment the hand from the rest of the image.
A common way is to define the hand region as skin pixels in the color space and to treat the rest as background pixels \cite{franzgrote2011palmprint,choras2012contactless,zhao2013sift,ungureanu2016palm}, or use methods like the Otsu threshold to binarize the image \cite{de2011approach,moco2014smartphone, chai2016palmprint}.
Then, hands would be aligned for the next step.
For this purpose, Franzgrote $\emph{et al.}$ in \cite{franzgrote2011palmprint} resorted to hand orientation normalization, while Zhao $\emph{et al.}$ in \cite{zhao2013sift} proposed a projection transformation model for estimating the matched SIFT feature points.
In \cite{li2012principal}, Li $\emph{et al.}$ extracted principal lines from palms and used them to align hands.
On the basis of hand segmentation and alignment, some schemes used valley detection \cite{ungureanu2017unconstrained}, Harris corner detection \cite{ungureanu2016palm} or radial distance functions \cite{aoyama2013contactless} to locate key points and established the local coordinate system for palmprint ROI extraction, while others determined key points based on hand contours \cite{fang2015mobile}.

There are also some ROI extraction pipelines without relying on hand segmentation as a pre-processing step. In \cite{han2007embedded}, Han $\emph{et al.}$ used a predefined preview frame on screen to align the position between hand and camera.
When the palm was aligned, they cropped a sub-image of fixed size at a fixed position as ROI.
Similar to Han \emph{et al.}'s work \cite{han2007embedded}, there are also methods such as hand-shaped guide window \cite{kim2015empirical}, double assistance points \cite{gao2018palmprint} and so on \cite{putra2014android,tiwari2016palmprint,leng2018palmprint} to guide the user's hand gestures.
Differently, Aykut $\emph{et al.}$ in \cite{aykut2015developing} used an advanced hand model learned by AAM to obtain key points at the palm contour. Later in \cite{zhang2017towards}, Zhang $\emph{et al.}$ used a binary palm mask to locate the finger-gap-points to construct a local coordinate system. However, Zhang \emph{et al.}'s method  \cite{zhang2017towards} can only be applied to images with uniform backgrounds.

\subsection{Approaches for Palmprint Verification}\label{subsect:palmprintVerification}
In the field of palmprint verification, great efforts have been devoted to extracting features from palmprint ROIs. Existing schemes roughly fall into five branches.

The first branch of palmprint feature extraction schemes are coding-based methods.
In a typical coding-based scheme, each field of the code map is assigned a bit-wised code, based on quantization of the image's response to a set of filters.
Conventional coding-based schemes include PalmCode \cite{zhang2003online}, CompCode \cite{kong2004competitive}, RLOC \cite{jia2008palmprint}, OLOF \cite{sun2005ordinal}, Accelerated CompCode \cite{franzgrote2011palmprint}, DRCC \cite{xu2016discriminative}, and CR-CompCode \cite{zhang2017towards}.
Recently, some more advanced coding schemes that attempted to obtain more direction information such as DDBC \cite{fei2019learning} and ALDC \cite{fei2019local}, have been proposed.

The second branch is the line-based branch.
These approaches typically extract palm lines using newly developed or off-the-shelf line detection operators.
Palm lines are then matched directly \cite{wu2006palm,huang2008palmprint} or represented in other forms \cite{zhang2004characterization,xu2014combining} for matching.

The third branch is the subspace-based branch.
These methods usually attempt to find a set of basis images from a training set and represent any probe image as a linear combination of them.
Typical subspace-based methods for palmprint applications include PCA \cite{cheung2006does}, CCA \cite{wang2013face}, LDA and ICA.
On the basis of these methods, more advanced methods have been put forward, LPDP \cite{gui2010locality}, Evo-KPCA \cite{ibrahim2014robust} and so on.
However, subspace-based methods are very sensitive to palmprint rotation, scaling and translation, and are not suitable for complex application scenarios such as mobile palmprint matching.

The fourth branch is the local-texture-descriptor-based branch.
The first kind of local texture descriptors are designed for general applications but are adapted to the field of palmprint verification, such as LBP, HOG \cite{jia2013histogram}, SIFT \cite{genovese2014touchless} and LDP.
In addition, local texture descriptors specially designed for palmprints have been proposed, and the typical ones include LBP+SIFT \cite{javidnia2015palm} and IHOL \cite{arunkumar2016palm}.

The fifth branch is the learning-based branch, which has attracted much attention in recent years.
Most approaches in this branch simply make use of off-the-shelf CNN structures as palmprint feature encoders \cite{jalali2015deformation,minaee2016palmprint,meraoumia2017improving,tarawneh2018pilot,ramachandra2018verifying,izadpanahkakhk2019joint,zhong2019centralized,matkowski2019palmprint}.
Only a few of them aimed to design new CNN architectures to embed the special traits of palmprint images \cite{svoboda2016palmprint,zhang2018bidirectional,zhou2019double,genovese2019palmnet}.

\subsection{Contactless Palmprint Datasets Publicly Available}\label{subsect:palmDataset}
To facilitate the design of palmprint matching algorithms, some researchers have collected benchmark contactless palmprint datasets and made them publicly available.

In \cite{hao2008multispectral}, Hao $\emph{et al.}$ collected a normal palm image dataset and a multispectral palm image dataset (CASIA dataset and CASIA-multispectral\cite{CASIA}, respectively).
The former contains 5,502 palm images from 312 subjects in a single session while the latter contains 7,200 palm images captured from 100 different people in 2 separate sessions.
For CASIA-multispectral, during each session, there are 3 samples and each sample includes 6 palm images captured under 6 different spectrum respectively. Furthermore, 4,800 images of CASIA-multispectral are palmprint images (taken under visible spectrums) and the other 2,400 ones are palm vein images (taken under IR spectrums).
Meanwhile, Kumar \cite{kumar2008incorporating} collected a contactless palmprint dataset (referred to as IIT-D \cite{IIT}), in which images were captured from 235 subjects in a single session using a self-developed device.
Later, Kanhangad $\emph{et al.}$ in \cite{kanhangad2010contactless} collected contactless 3D and 2D hand images to build the PolyU palmprint dataset \uppercase\expandafter{\romannumeral2} \cite{PolyU}.
Kanhangad \emph{et al.}'s dataset contains right-hand images from 114 subjects in 5 different poses.
In \cite{ferrer2011bispectral}, Ferrer $\emph{et al.}$ released a contactless palmprint dataset called GPDS100 \cite{GPDS100}.
GPDS100 comprises 2,000 samples that were acquired from 100 palms over two sessions.
For each palm, GPDS100 kept 10 images.
The dataset collected by Aykut $\emph{et al.}$ in \cite{aykut2015developing} is referred to as KTU dataset \cite{KTU}, which comprises 1,752 images collected from 145 different palms in a single session.
Concurrently, Hassanat $\emph{et al.}$ in \cite{hassanat2015new} collected the MOHI dataset and WEHI dataset \cite{MOHI&WEHI} using mobile phone cameras and webcams, respectively. More recently, Joshi $\emph{et al.}$ collected 1,344 high-resolution palmprint images from 168 subjects to compose the COEP palmprint database \cite{COEP}.
Zhang $\emph{et al.}$ \cite{zhang2017towards} collected 12,000 palm images from 600 palms in two different sessions using their established contactless palmprint collection device, which is referred to as Tongji contactless palmprint dataset \cite{TongjiPD}.

\subsection{Our Motivations and Contributions}\label{subsect:motivationContri}
Through the study of relevant literature, we find that there is still much room for improvement in the field of palmprint verification for smartphones in at least two aspects.

First, large-scale datasets are quite rare in this field.
To design palmprint verification systems for smartphones and also to objectively compare their performance, a public large-scale benchmark dataset with various backgrounds and illumination conditions, collected by smartphones and  carefully labeled, is indispensable.
Unfortunately, thus far, the research community lacks such a dataset.

Second, effectively and accurately matching palmprints on smartphones remains a challenge.
Due to the hardware limitation of smartphones and the speed requirements of mobile scenes, palmprint recognition on smartphones needs to meet both the requirements of light weight and fast speed.
The performance of palmprint verification systems on smartphones still have much room for improvement, mainly regarding two tasks: palmprint ROI extraction and ROI matching.
\begin{enumerate}
\item{Because of the complicated backgrounds and illumination conditions of palmprint images collected in uncontrolled environments, it is a challenging task to delineate palmprint ROIs accurately.
In order to get rid of the influence of complex backgrounds in palmprint images, existing palmprint verification systems on smartphone usually need to segment hand at first, or request the user to place the hand in the designated position in front of the camera. The former is time-consuming and error-prone, while the latter is not user-friendly.
Moreover, existing palmprint ROI extraction schemes hardly have the properties of scale and rotation invariance.
In short, when the position of the hand is not ideal or the background is too complex, the performance of these methods is often unsatisfactory due to their inherent limitations.}
\item{In the phase of palmprint ROI matching, it is hard to discriminate whether two palmprint ROIs belong to the same hand due to the position deviation between the two ROIs or the difference of palmprint postures.}
\end{enumerate}

In this work, we attempt to fill the aforementioned research gaps to some extent. Our contributions are briefly summarized as follows:
\begin{enumerate}
\item{To facilitate the study of palmprint verification on smartphones, we have established a large-scale palmprint dataset named MPD and have made it online available.
MPD comprises 16,000 palm images collected from multi-brand smartphones and all the images are manually labeled with care. To our knowledge, it is currently the largest publicly available contactless palmprint dataset collected by smartphones in the unconstrained environment.
Such a dataset will benefit the studies of palmprint recognition algorithms on smartphones.
Please refer to Sect. \ref{sect:MPD} for more details about MPD.
}
\item{
We proposed a data-driven learning-based system named DeepMPV+ to verify palmprints on smartphones.
Specifically, DeepMPV+ consists of two main modules, the ROI extraction module and the ROI matching module.
Unlike the strategies reported in the literature, we creatively model the ROI extraction problem as an object detection problem, which can be naturally and effectively solved by modern CNNs.
For these reasons, the proposed ROI extractor has the merits of being fast, accurate, and highly robust to complex backgrounds.
In the stage of ROI matching, due to the ability of ArcFace \cite{deng2019arcface} to make different kinds of feature vectors more evenly and more independently located in the super feature space, and considering the hardware limitations and speed requirements of smartphones, a palmprint verifier based on MobileFaceNet \cite{chen2018mobilefacenets} and ArcFace is used to match the extracted palmprint ROI.
DeepMPV+ can work well on smartphones, and can match palmprint images with slight position offset and gesture deflection.
The efficiency and performance of DeepMPV+ have been thoroughly evaluated in experiments.
}
\item{
Using the developed algorithms, we have implemented a practical palmprint verification system for the iPhone 8 Plus platform.
As far as we know, it is the first reported smartphone-oriented palmprint verification system.
To make our results fully reproducible, our collected dataset MPD and all the relevant code have been publicly released at \url{https://cslinzhang.github.io/MobilePalmPrint/.}
Our research will prompt people to realize that palmprint is also a viable option for identity authentication on smartphones besides fingerprint and face.
}
\end{enumerate}

A preliminary version of this manuscript has been accepted by ICME 2019 \cite{zhang2019pay}.
The following improvements are made in this version: 1) the whole pipeline is simplified from DeepMPV to DeepMPV+ with higher efficiency and performance, using 2 DCNN models instead of 3; 2) a thorough performance evaluation of modern DCNN models in the context of palmprint verification on smartphones is conducted, and higher-performance lightweight models are selected based on the experimental results; 3) the performance of palmprint recognition models and metrics used to measure the quality of palmprint recognition is thoroughly investigated and analyzed; and 4) additional relevant approaches are evaluated in experiments.

\section{DeepMPV+: A Learning-Based Pipeline for Palmprint Verification}\label{sect:deepMPV}
\begin{figure}
\centering
\includegraphics[width=1.0\columnwidth]{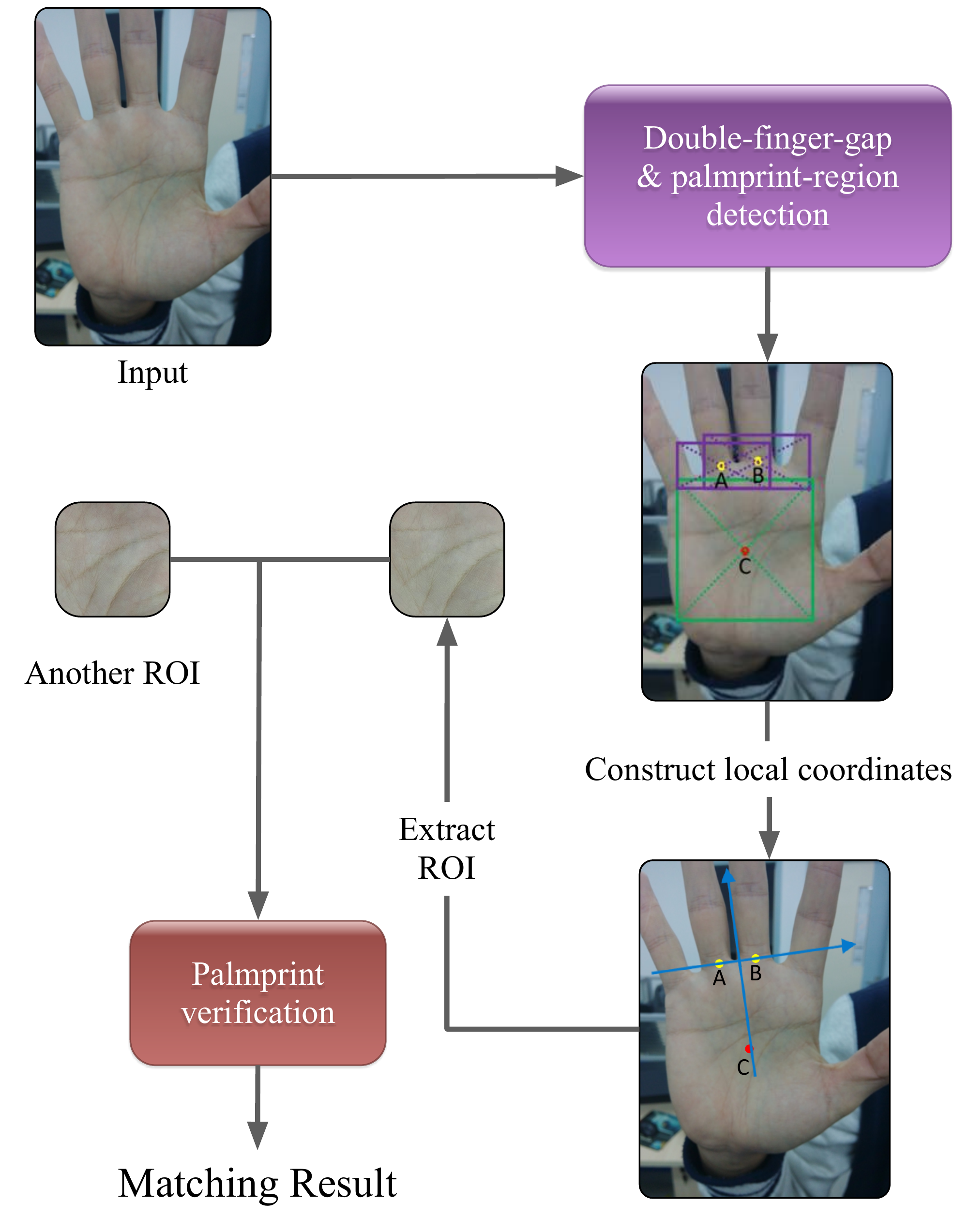}
\caption{The system flowchart of $DeepMPV$.}
\label{fig:1}
\vspace {-0.25cm}
\end{figure}

In this section, the proposed smartphone-oriented palmprint verification system DeepMPV+ is presented in detail. DeepMPV+ consists of two key components (as shown in Fig. \ref{fig:1}), including palmprint ROI extraction and ROI verification. The details will be introduced in the following subsections.

\subsection{Palmprint Image Annotation}

\begin{figure}
  \centering
  \subfigure[]{
   \label{fig:2a}
   \includegraphics[width=2.5in]{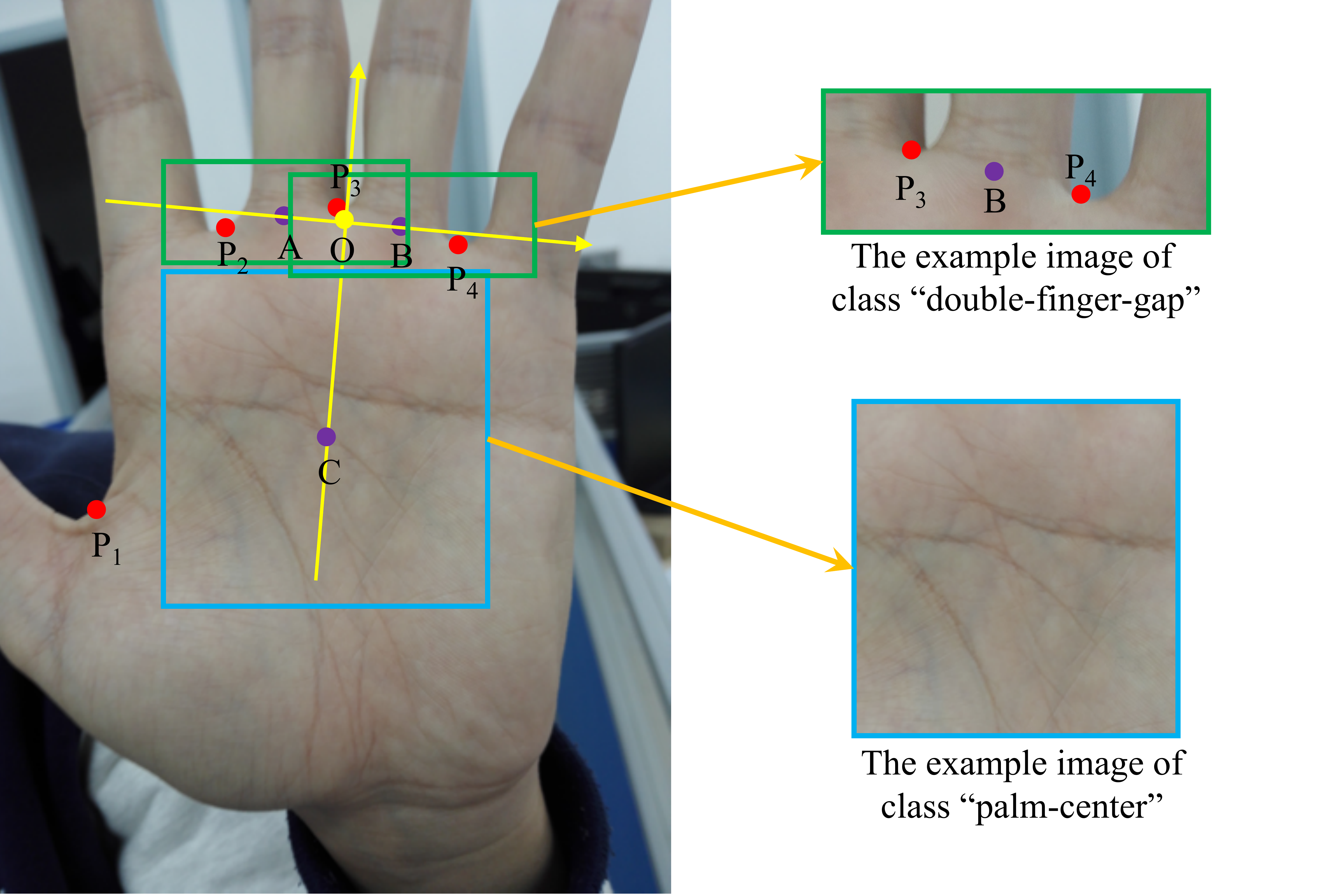}
   }
   \hspace{1in}
   \subfigure[]{
   \label{fig:2b}
   \includegraphics[width=2.5in]{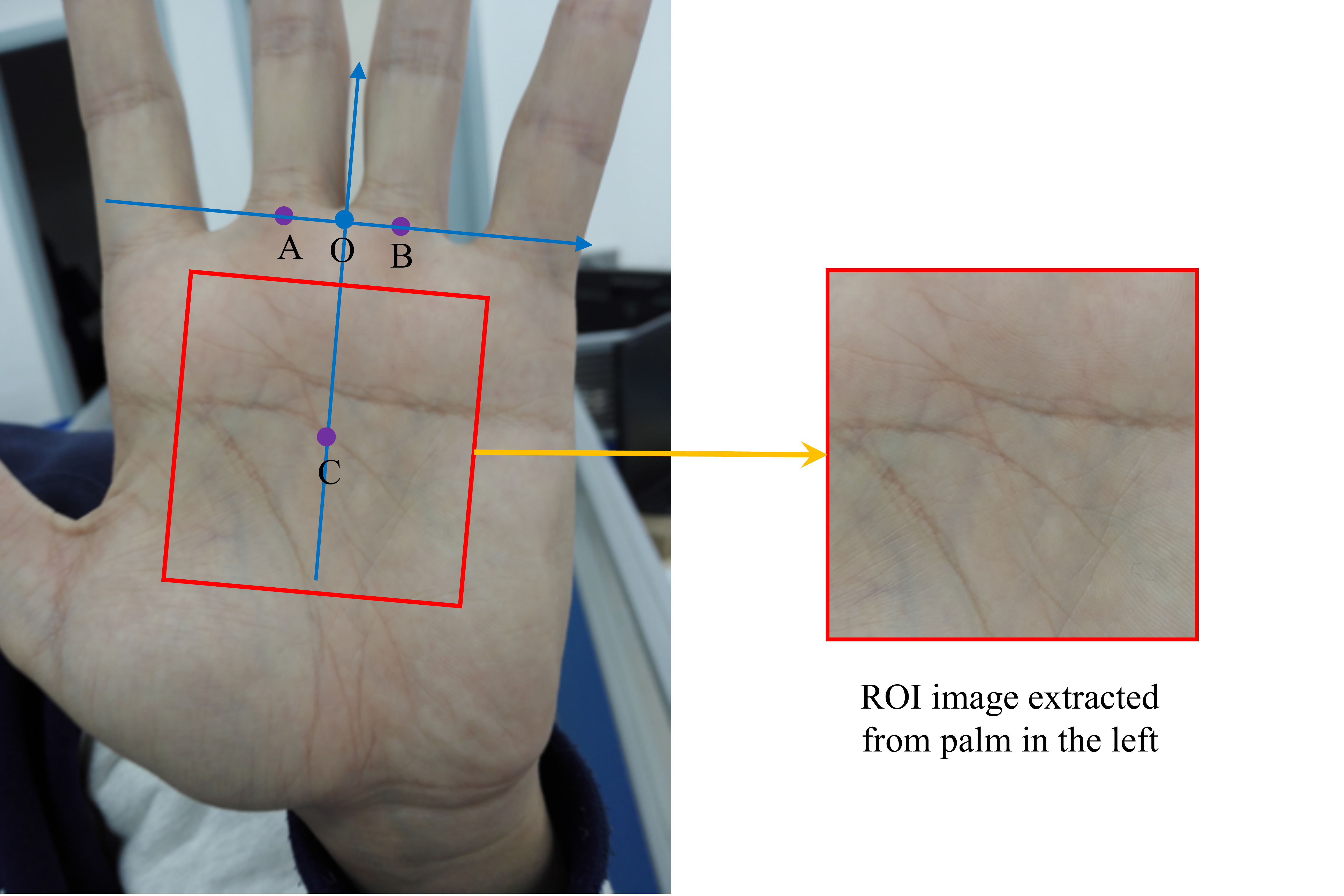}
   }
   \caption{Illustration of palm image annotation. (a) In the data annotation stage, we mark finger-gap-points (red points); the green rectangles indicate the double-finger-gaps while the blue rectangle represents the palm-center. (b) We use the well-trained detector $\textbf{\emph{D}}$ to obtain the position of point (A, B, C) and construct a local coordinate system (marked with blue arrows) to extract the $ROI$ (marked as a red square).}
   \label{fig:2} 
\end{figure}

When DCNNs are applied in the field of palmprint, a large-scale and carefully labeled palmprint image dataset is indispensable. First, the ground-truth of the object detection needs to be determined. For a palm image, the finger-gap-points (points at the junction of adjacent fingers, denoted as $\rm P_{i}$ and marked by the red points in Fig. \ref{fig:2a}) between fingers are the most obvious visual feature areas. However, a local coordinate system cannot be established only by these disorderly finger-gap-points. Therefore, finger-gap-points cannot be directly used as regression objects for object detection.

To determine a local coordinate system, at least three points that are not on the same line need to be found. The first two points A and B of these three points are determined from the aforementioned finger-gap-points. Point A is the midpoint of $\rm P_{2}P_{3}$ while point B is the midpoint of $\rm P_{3}P_{4}$. For these finger-gap-points, we name the point between the thumb and index finger as thumb-gap-point (the red point $\rm P_{1}$ in Fig. \ref{fig:2a}), while the other finger-gap-points are named as general-finger-gap-point (the red points $\rm P_{2}\sim P_{4}$ in Fig. \ref{fig:2a}). To eliminate the interference of thumb-gap-points on detection, the double-finger-gap including one general-finger-gap-point pair (($\rm P_{2}$, $\rm P_{3}$) or ($\rm P_{3}$, $\rm P_{4}$)) is regarded as the first class of object detection. The double-finger-gaps are marked with green rectangles in Fig. \ref{fig:2a}. Point A and point B are also the center points of two double-finger-gaps respectively.
Based on them, the X-axis of the local coordinate system can be determined.

After determining point A and point B, point C is still needed to uniquely determine the local coordinate system. The midpoint O of AB is defined as the origin of the local coordinate system, and point C is defined as the point such that $\rm OC = \frac{3}{2} \times \left \| {AB} \right \|$ that is located on both the Y-axis and the palm center area.
The palm-center with point $\rm C$ as the center, containing the main palmprint area, is regarded as the second class of object detection, and is marked by the blue rectangle in Fig. \ref{fig:2a}.

In the stage of image annotation, all the finger-gap-points {$\rm P_{i}$} are carefully labeled.
We designate $\overrightarrow{\rm AB}$ as the positive direction of the X-axis, establish the local coordinate system (marked with yellow arrows in Fig. \ref{fig:2a}), and then calculate the position and area of the ``double-finger-gap'' and the ``palm-center'' based on the aforementioned finger-gap-points. This position and area information will be used as the ground truth of the object detection model. Examples of the double-finger-gap and the palm-center are provided in Fig. \ref{fig:2a}.

\subsection{Palmprint ROI Extraction}
After data annotation, the next step is to train the object detection model. Having investigated the literature, we find that Tiny-YOLOV3 \cite{yolov3} is a state-of-the-art general-purpose object detector based on DCNN for mobile devices. Compared with MobileNetV1+SSD  \cite{howard2017mobilenets,liu2016ssd}, PeleeNet \cite{wang2018pelee} and other lightweight networks, Tiny-YOLOV3 has higher speed and accuracy in real-time object detection on mobile devices. Hence, our detector $\textbf{\emph{D}}$ is based on Tiny-YOLOV3.

Given a palm image, the double-finger-gap is detected to obtain the coordinates of point pair (A, B), and the palm-center is detected to obtain the coordinates of point C. As shown in Fig. \ref{fig:2b}, the line through $\rm AB$ is defined as the line where the X-axis is located, while the direction perpendicular to $\rm AB$ and opposite to point C is defined as the positive direction of the Y-axis. The direction of the 90-degrees clockwise rotation of the Y-axis is the positive direction of the X-axis. With the midpoint O of $\rm AB$ as the origin, the local coordinate system for a given image is successfully established, which is marked with blue arrows in Fig. \ref{fig:2b}.

After the establishment of the local coordinate system, $ROI$ is defined as a square area parallel to both the X-axis and the Y-axis. Suppose the coordinate of point O is ($\rm x_{0}$, $\rm y_{0}$), the center point of the $ROI$ is set as ($\rm x_{0}$, $\rm y_{0}+\frac{3}{2} \times \left \| {AB} \right \|$), and its side length is set as $\rm s_{R}$. The side length $\rm s_{R}$ is determined as $\rm s_{R} = \frac{5}{2} \times \left \| {AB} \right \|$. Finally, we can extract the palmprint ROI from the palm image by extracting the image region covered by the $ROI$ (marked as red square in Fig. \ref{fig:2b}). The pipeline of palmprint ROI extraction is summarized in Algorithm \ref{alg:pipelined}.

\begin{algorithm}[H]
\caption{Overall Pipeline of Palmprint ROI Extraction}
\label{alg:pipelined}
\begin{algorithmic}[1]
  \REQUIRE
  A palm image $\textbf{I}$ and our detector $\textbf{\emph{D}}$.
  \ENSURE
  Palmprint ROI image $ROI$.
\STATE Feed $\textbf{I}$ to $\textbf{\emph{D}}$ to obtain double-finger-gap center point set $\rm P_{i}^{D}$ and the palm-center center point C, where the size of set $\rm P_{i}^{D}$ is $L$;
\IF {$L>2$}
    \STATE Traverse the set $\rm P_{i}^{D}$, selecting point pairs;
    \STATE Calculate the distance between two points;
    \STATE Choose the two farthest points as (A,B);
\ELSIF {$L<2$}
    \RETURN an indication of incomplete detection
\ENDIF
\STATE Set the line through AB as the X-axis;
\STATE Set the direction perpendicular to AB and opposite to point C as the positive direction of the Y-axis;
\STATE Set the midpoint of AB as the origin ($\rm x_{0}$, $\rm y_{0}$);
\STATE Take $\rm \left \| {AB} \right \|$ as the unit length of axis;
\STATE Construct the local coordinate system;
\STATE Set the center coordinate of the $ROI$ to ($\rm x_{0}$, $\rm y_{0} + \frac{3}{2} \times \left \| {AB} \right \|$);
\STATE Set the side length of $ROI$ to $\rm \frac{5}{2} \times \left \| {AB} \right \|$;
\RETURN segmented $ROI$
\end{algorithmic}
\end{algorithm}

The details of $\textbf{\emph{D}}$'s training set are shown in Sect. \ref{sect:MPD}. In Sect. \ref{sect:Experiments}, we will quantitatively evaluate the performance of the detector $\textbf{\emph{D}}$.

\subsection{Palmprint Verification}

For palmprint verification, the input is a pair of palmprint ROIs. In traditional methods, palmprint features are usually extracted to calculate matching scores. Therefore, the method used to extract palmprint features matters greatly. To extract the features of ROIs better, we used a DCNN-based method to fulfill the task. By studying the literature, we found that ArcFace\cite{deng2019arcface} is very suitable for palmprint verification.

ArcFace uses an Additive Angular Margin Loss to obtain highly discriminative features from faces. It can make the feature vectors of different classes more uniform and independent in the super-feature space and help the model converge better. Moreover, considering the hardware limitations of the smartphone platform, after a detailed experiment and comparison based on the experimental results, we chose to use MobileFaceNet \cite{chen2018mobilefacenets} as the backbone to extract palmprint features for matching. MobileFaceNet is a lightweight network designed for face verification and is an improved version of MobileNetV2 \cite{sandler2018inverted}. On the basis of MobileNetV2, in addition to the Additive Angular Margin Loss, it also used a Global Depthwise Convolution (GDConv) instead of Average Pooling layer, so that the network could treat the center point and the corner point differently when updating the weight. This model is very suitable for palmprint verification on smartphones, which is also an important reason for us to choose it. Therefore, our verifier $\textbf{\emph{V}}$ is based on MobileFaceNet and ArcFace.

Specifically, take a $224\times224$ palm ROI as the input. The $512 \times 1$ vector output by $\textbf{\emph{V}}$ is taken as the feature of this palm ROI. The final matching score can be obtained by performing a dot multiplication operation on the two palmprint ROI features. The threshold value to determine whether the match is successful is set to $T$, and the value of $T$ is determined according to the experiments in Sect. \ref{subsect:performMbfn}.
We took scores higher than $T$ as matching success, otherwise, the palmprints matching fails.
Details of the palmprint verification pipeline are shown in Algorithm \ref{alg:pipelinev}.

\begin{algorithm}[t]
  \caption{Overall Pipeline of Palmprint ROI Verification}
  \label{alg:pipelinev}
  \begin{algorithmic}[1]
    \REQUIRE
    A pair of palmprint ROI images $ROI_{1}$ and $ROI_{2}$, our verifier $\textbf{\emph{V}}$, and our threshold $T$.
    \ENSURE
    Matching result $R$.
    \STATE Resize $ROI_{1}$ and $ROI_{2}$ to the size $224\times224$;
    \STATE Feed $ROI_{1}$ and $ROI_{2}$ to $\textbf{\emph{V}}$ to obtain palmprint features $F_{1}$ and $F_{2}$, respectively;
    \STATE Use $l_2$-norm to normalize features $F_{1}$ and $F_{2}$ to obtain features $Fnorm_{1}$ and $Fnorm_{2}$, respectively
    \STATE Calculate matching score $S$ by $S = Fnorm_{1} \cdot Fnorm_{2}$;
    \IF {$S>=T$}
        \STATE Set matching result $R$ as ``Matching Success'';
    \ELSIF {$S<T$}
        \STATE Set matching result $R$ as ``Matching Fail'';
    \ENDIF
    \RETURN matching result $R$.
  \end{algorithmic}
\end{algorithm}

More details of verifier $\textbf{\emph{V}}$'s training set will be explained in Sect. \ref{sect:MPD}. In Sect. \ref{sect:Experiments}, its performance will be quantitatively evaluated.

\section{MPD: A Benchmark Dataset for Mobile Palmprint Verification}\label{sect:MPD}

\begin{figure*}
  \centering
  \subfigure[]{
   \label{fig:3a}
   \includegraphics[width=1.2in]{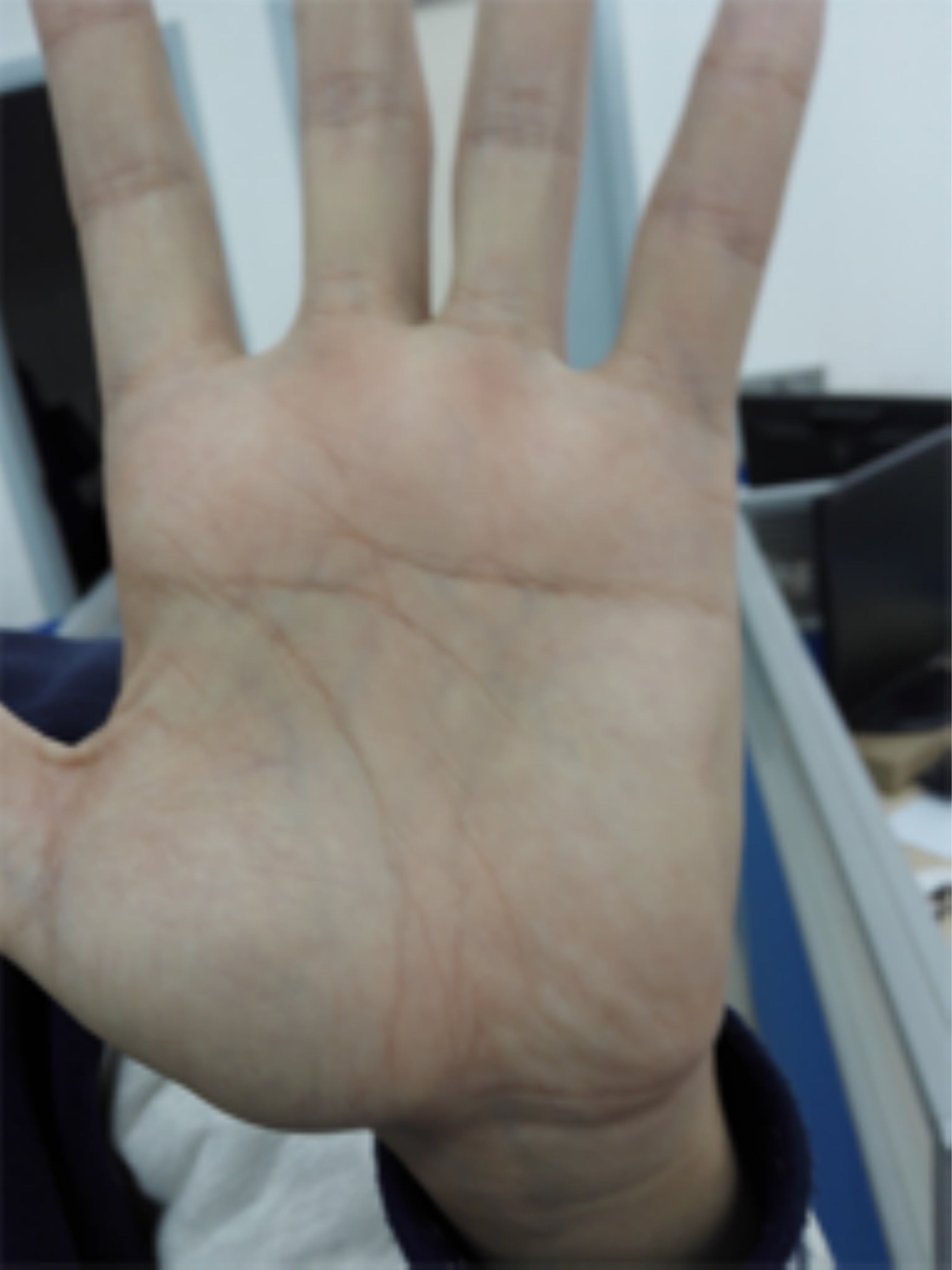}
   }
   \subfigure[]{
   \label{fig:3b}
   \includegraphics[width=1.2in]{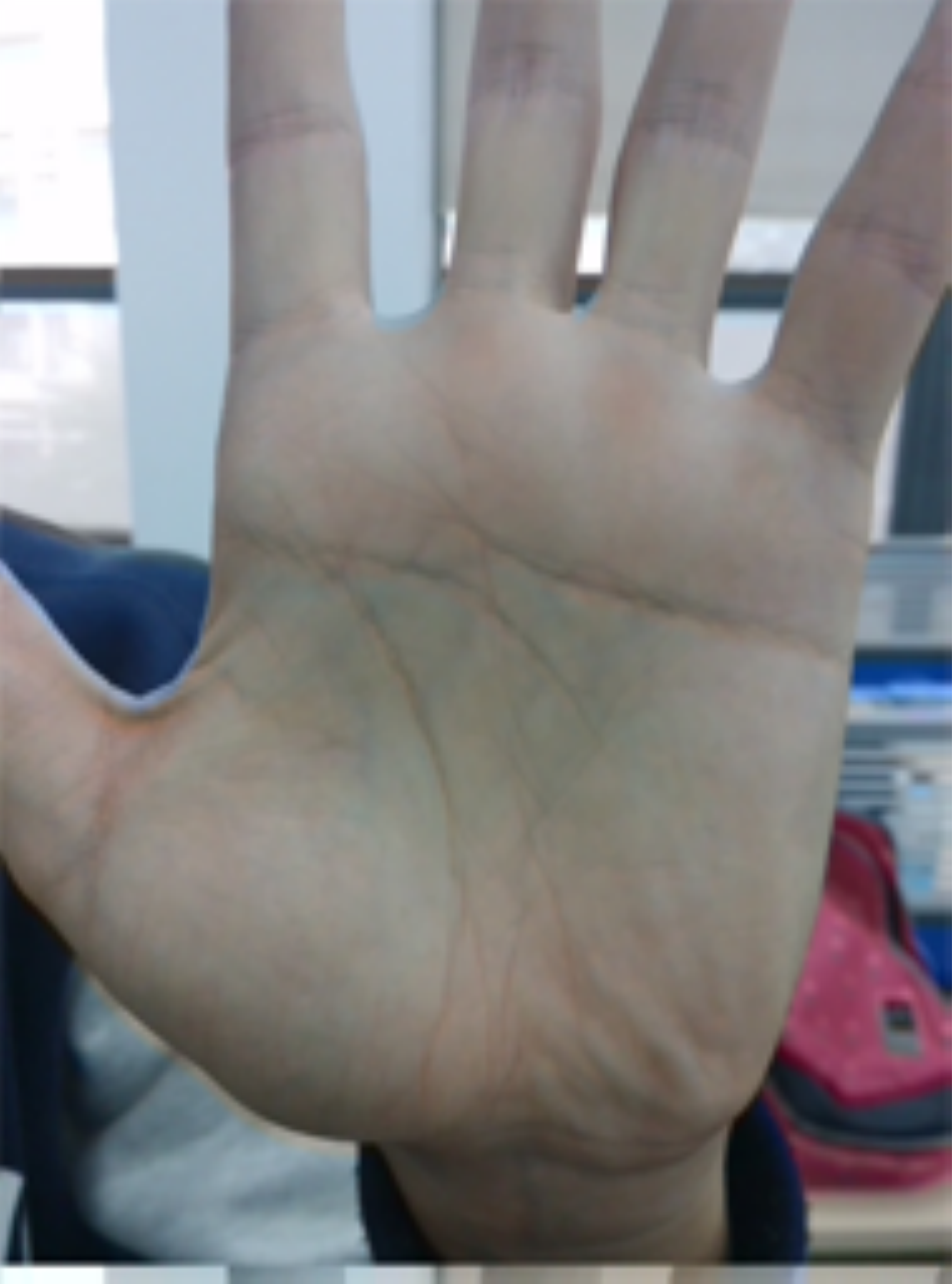}
   }
   \subfigure[]{
    \label{fig:3c}
    \includegraphics[width=1.2in]{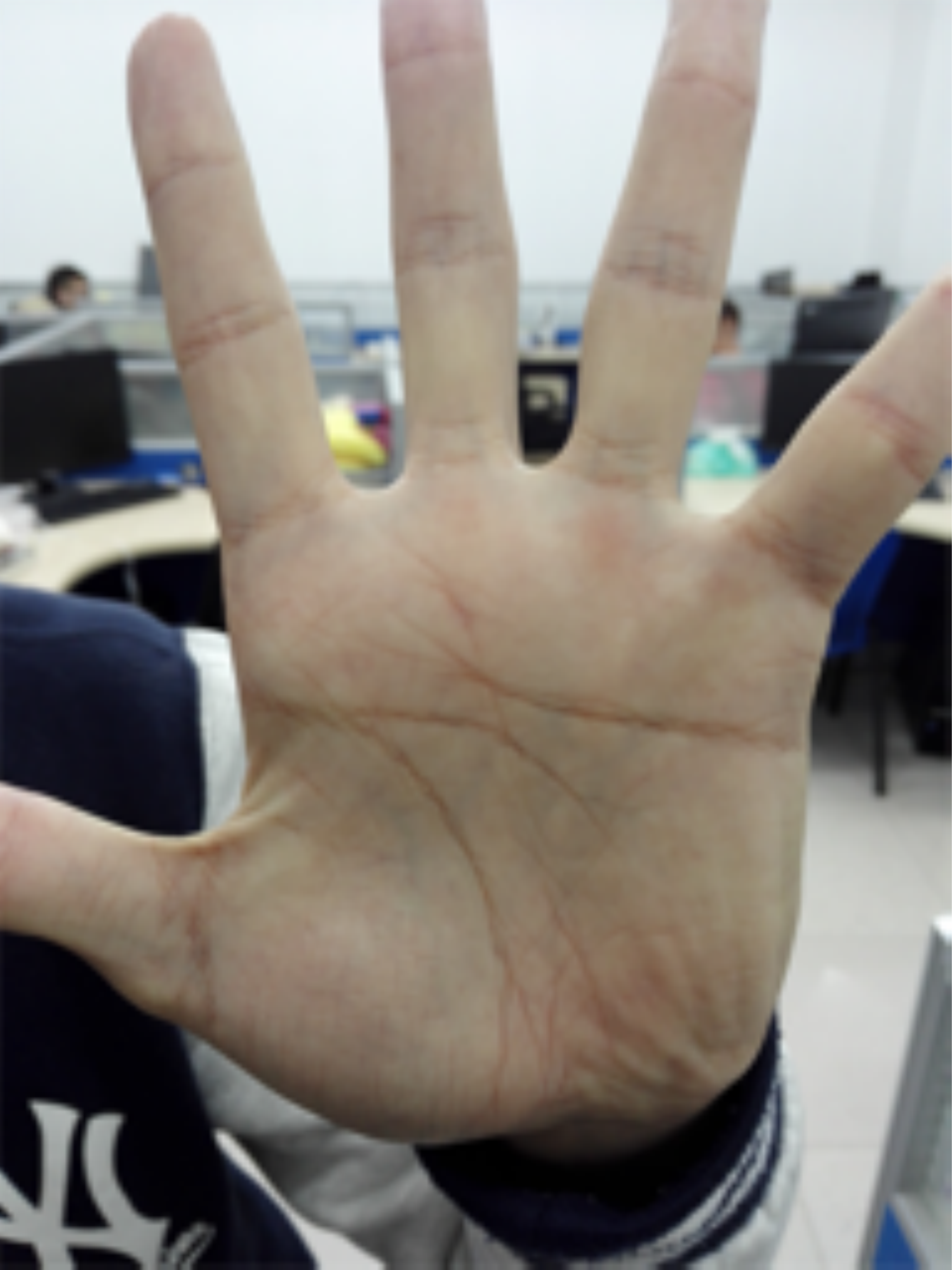}
    }
    \subfigure[]{
    \label{fig:3d}
    \includegraphics[width=1.2in]{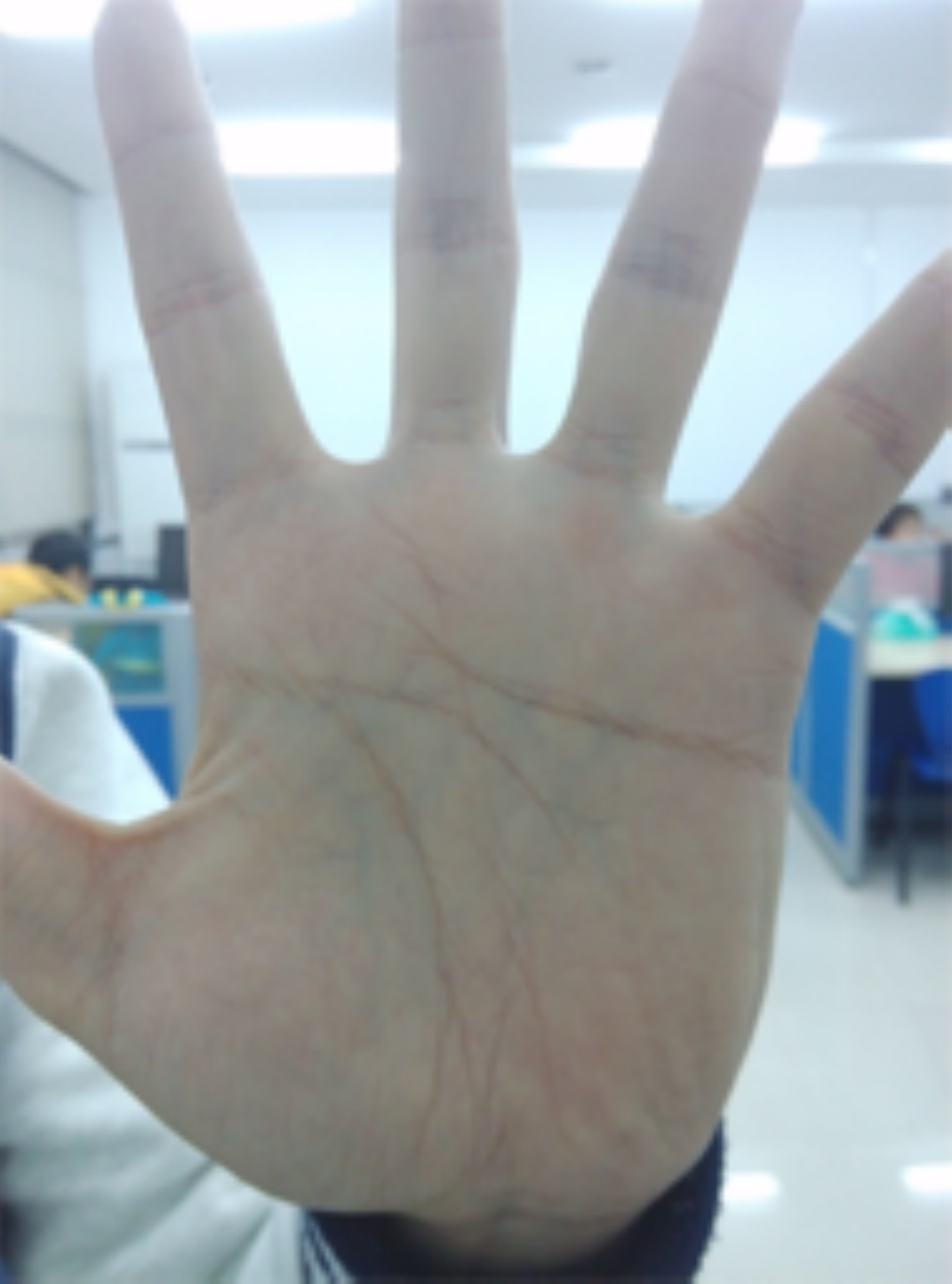}
    }
   \hspace{1in}
   \subfigure[]{
    \label{fig:3e}
    \includegraphics[width=1.2in]{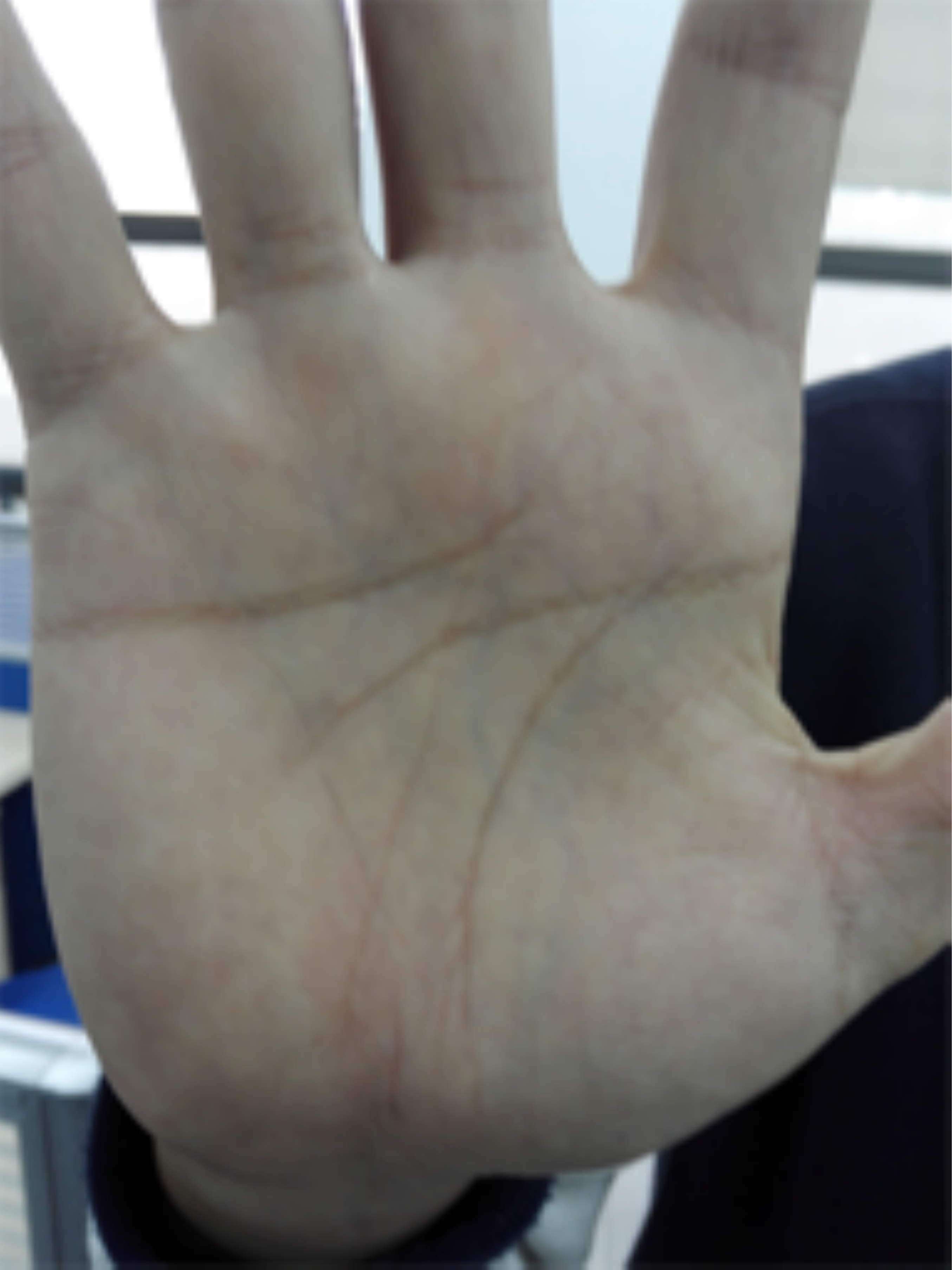}
    }
    \subfigure[]{
    \label{fig:3f}
    \includegraphics[width=1.2in]{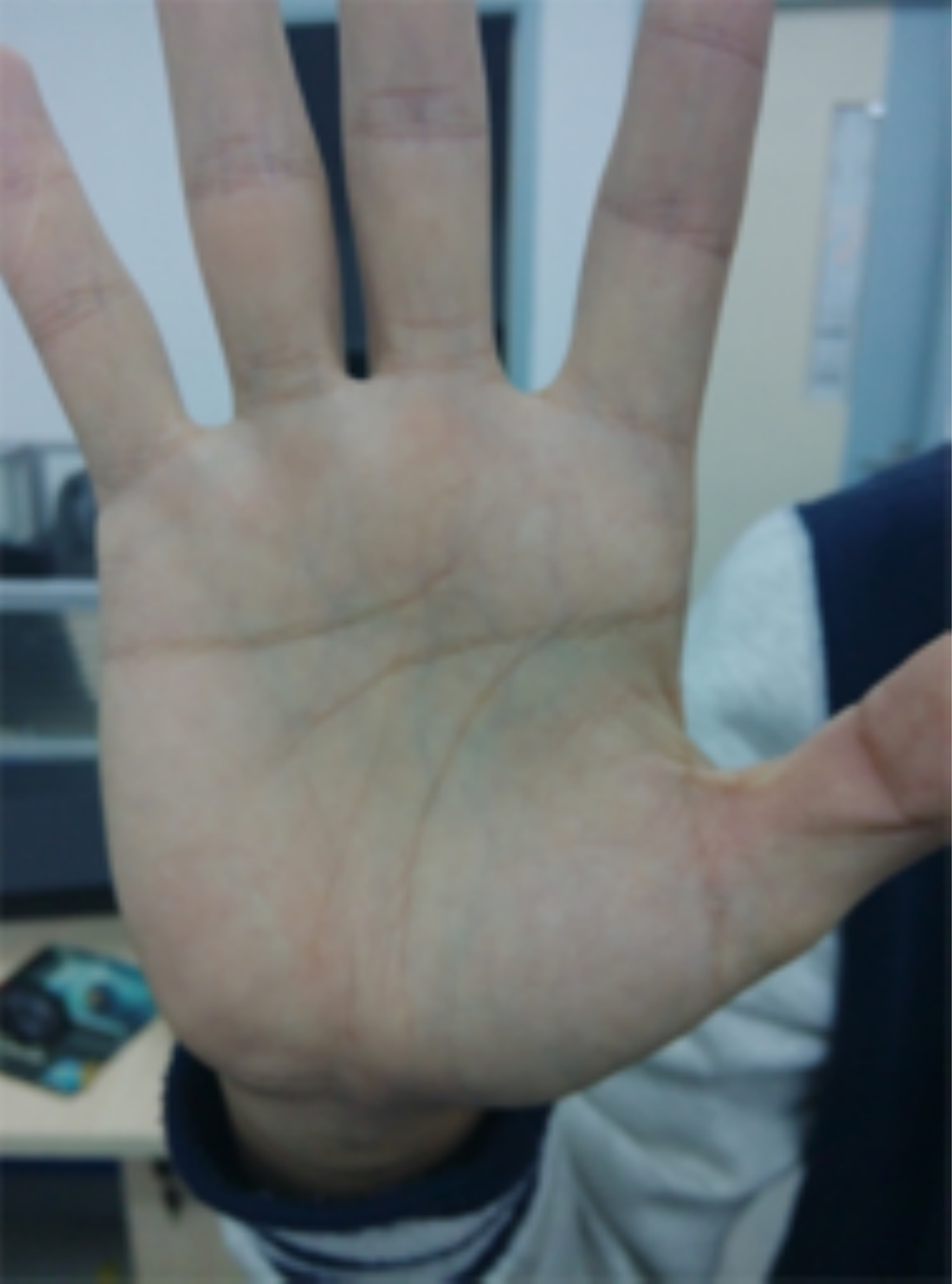}
    }
    \subfigure[]{
     \label{fig:3g}
     \includegraphics[width=1.2in]{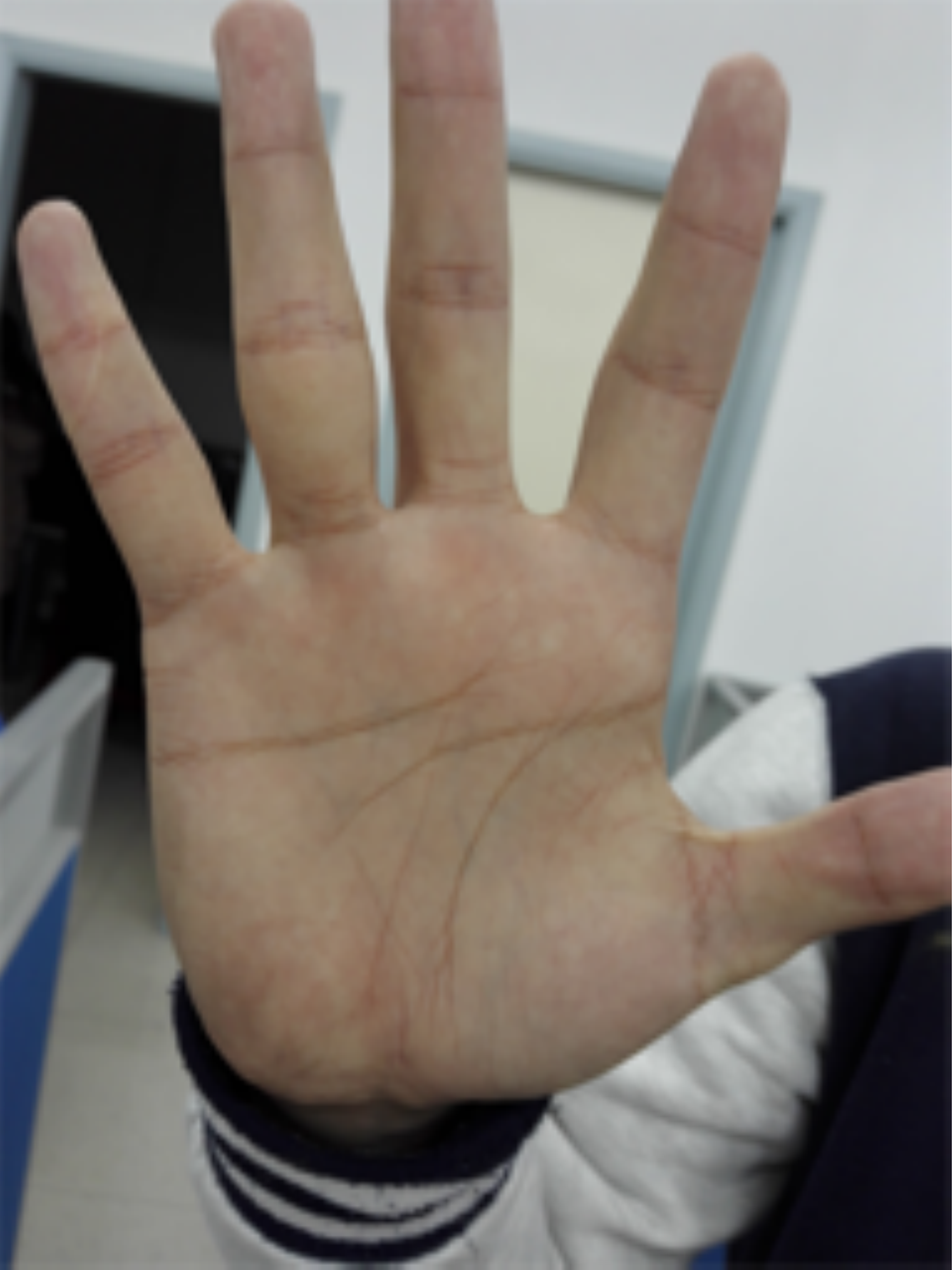}
     }
     \subfigure[]{
     \label{fig:3h}
     \includegraphics[width=1.2in]{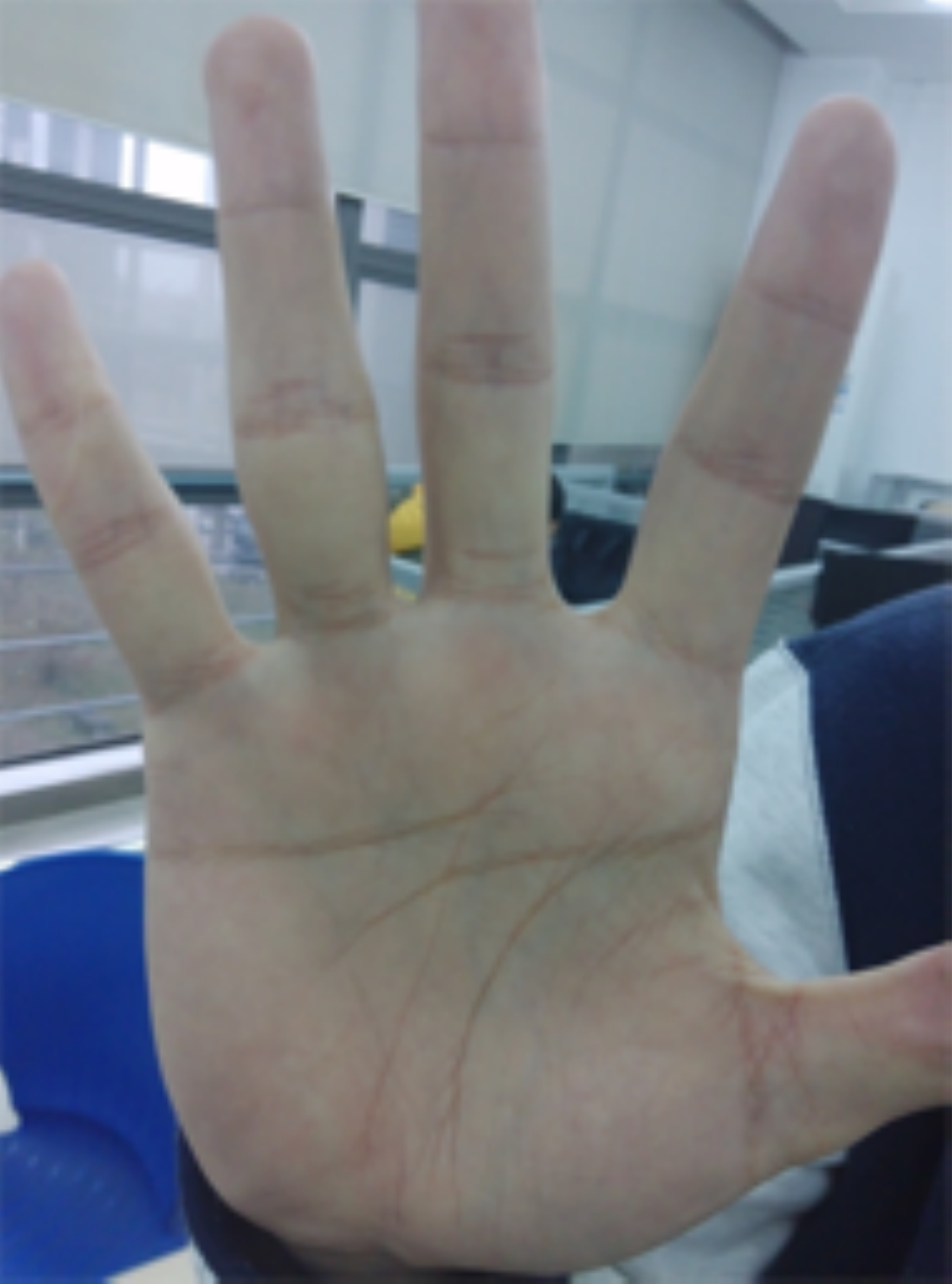}
     }
  \caption{(a)$\sim$(h) are 8 images from the same person in our palmprint dataset MPD. (a)$\sim$(d) are left hand and (e)$\sim$(h) are the right hand. (a), (c), (e) and (g) were captured by Huawei, and (b), (d), (f) and (h) were captured by Xiaomi. (a), (b), (e) and (f) were collected in the first session, and (c), (d), (g) and (h) were collected in the second session.}
  \label{fig:3}
\end{figure*}

As stated in Sect. \ref{sect:RelatedWorkContri}, in view of the fact that a database specially dedicated to mobile palmprint verification is still lacked in the community, and to provide a reasonable performance evaluation benchmark for palmprint verification on mobile platforms, we are motivated to establish such a dataset in this work. In this section, the establishment of our mobile palmprint database MPD is described. In total, MPD has 16,000 palmprint images from 200 subjects, and for each image there is an associated annotation with the position of finger-gap-points. The construction of MPD comprises two main steps: palm image collection and training and test set preparation. Details are given in the following subsections.

\subsection{Palm Image Collection}
To provide a reasonable performance evaluation benchmark for palmprint verification on mobile platforms, a large number of palm images were collected to compose our dataset MPD. MPD contains palm images with a variety of backgrounds and lighting environments. For the purpose of eliminating the influence of the camera parameters of different brands of mobile devices, two kinds of smartphones were used to collect palm photos: Huawei and Xiaomi. To avoid the influence of the season or time on photographs, after the first round of collection, we took a group of photos with the same two mobile phones of the same group of people according to the same standard half a year later. MPD comprises 16,000 palmprint images from 200 volunteers. Those volunteers were staffs or students of Tongji University, with a balanced gender ratio. Among them, 195 subjects were 20 $\sim$ 30 years old and the others were 30 $\sim$ 50 years old. We obtained 10 photos of each subject's hand with each smartphone in each time period, which means there are 40 images of each hand and we have 400 different hands in our dataset. Furthermore, we labeled all 16,000 palm images in our dataset.

The palmprint images are named according to the following rules: 1) The first three digits, taking ``001'' as an example, indicate that this palm image was taken from the volunteer numbered 1; 2) the fifth digit, ``1'' for example, means that this palm image was taken in the first period; 3) for the seventh digit, ``h'' means that this palm image was taken by Huawei and ``m'' means this photo was taken by Xiaomi; 4) for the ninth digit, ``l'' indicates that this palm image was taken from the left hand of the volunteer and ``r'' indicates that this image was taken from the right hand; 5) the last two digits show the number of the palm image taken by the same volunteer with the same hand at the same time using the same smartphone. For instance, a palm image named ``006\_2\_h\_r\_08.jpg'' is a picture taken from the right hand of our sixth volunteer (No. 6) using Huawei mobile phone in the second period. Examples of palm images are shown in Fig. \ref{fig:3}.

\subsection{Training and Test Set Preparation}

\begin{figure}
  \centering
  \subfigure[]{
   \label{fig:4a}
   \includegraphics[width=1.25in]{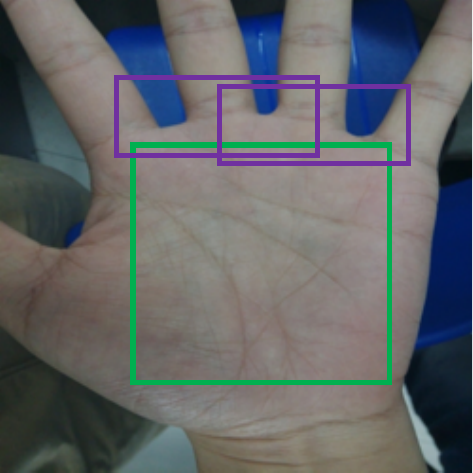}
   }
   \subfigure[]{
   \label{fig:4b}
   \includegraphics[width=1.25in]{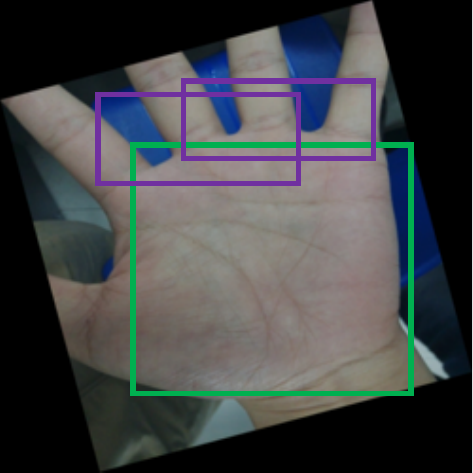}
   }
\caption{In the phase of preparing training samples, to make the detector $\textbf{\emph{D}}$ rotation invariant, each original labeled image was rotated to generate a set of its rotated versions. (a) is the original labeled image, and (b) is its rotated version generated from (a). The purple rectangles indicate the ``double-finger-gap'', and the green rectangles indicate the ``palm-center''.}
\label{fig:4}
\vspace {-0.25cm}
\end{figure}

In this section, the preprocessing and partitioning of each dataset will be introduced in detail. In dataset partitioning, palmprint detection and palmprint matching are regarded as open-set problems, and training set, verification set, and test set are guaranteed to have no overlap.

\subsubsection{Generation and Preprocessing of $\textbf{D}$'s Sets}
In the training stage for our detector $\textbf{\emph{D}}$, a training set was generated from the original dataset with annotations. The original dataset has annotations including the coordinates of finger-gap-points (general-finger-gap-point and thumb-gap-point). Subsequently, the general-finger-gap-points are used to calculate the positions and areas of the ``double-finger-gap'' and ``palm-center''. As we can see in Fig. \ref{fig:4}, the purple rectangles indicate the area of the ``double-finger-gap'', and the green square indicates the area of the ``palm-center''. For each palm image in our dataset, there are two marks of the class ``double-finger-gap'' and one mark of the class ``palm-center'' in general.

After generating the labels of our detector $\textbf{\emph{D}}$'s training set, to make it scale invariant, the training set was augmented by rotating each original labeled image to generate a number of its rotated versions as shown in Fig. \ref{fig:4}. In detail, for a given original labeled image $\textbf{I}$, $J$ rotated versions $\{\textbf{I}_{j}\}_{j=0}^{J-1}$ were obtained from it with a rotation angle $\theta_{J} = \frac{360}{J}$. To make the operation of rotation easier, the original labeled images were resized to a fixed size $s_{f} \times s_{f}$. Fig. \ref{fig:4a} is an original labeled image, and Fig. \ref{fig:4b} is its rotated version generated by Fig. \ref{fig:4a} with a rotation of 15 degrees.

The augmented dataset was separated according to the following ratio: training set:validation set:test set = 8:1:1. The final training and test sets of detector $\textbf{\emph{D}}$ were named $\rm{MPD\textsubscript{D}}$ (short for ``The Dataset for Detector Generated from Mobile Palmprint Dataset'').

\subsubsection{Generation and Preprocessing of $\textbf{V}$'s sets}
In the training stage for $\textbf{\emph{V}}$, based on labeled data, the ROI dataset was generated with 16,000 images from 400 hands of 200 persons. It was named $\rm{MPD\textsubscript{ROI}}$ (short for ``The ROI Dataset Generated from Mobile Palmprint Dataset'').

Since palmprint images acquired by smartphones often have different contrast, sharpness, and alignment quality, to improve the generalization ability of our verifier $\textbf{\emph{V}}$, the following operations are performed for data augmentation. Two different models $\emph{V\textsubscript{IR}}$ and $\emph{V\textsubscript{MBFN}}$ have been applied with different data augmentation. And the details of these models are described in Sect. \ref{subsect:verifierDetails}.
\begin{enumerate}
\item{For $\emph{V\textsubscript{IR}}$ in CASIA, (a) Resize to $246\times246$ and then random center crop, (b) ColorJintter: brightness and contrast are both $0.25$, saturation is $0.2$ and hue is $0.15$ and (c) Horizontal Flip. For $\emph{V\textsubscript{MBFN}}$ in CASIA, (a) Resize to $246\times246$ and then random center crop, (b) RandomResizedCrop: scale and ratio are both $[0.9, 1.1]$, (c) ColorJintter: brightness and saturation are both $0.25$, contrast and hue are both $0.15$ and (d) Horizontal Flip.}
\item{For $\emph{V\textsubscript{IR}}$ in IITD, (a) RandomResizedCrop: scale and ratio are both $[0.75, 1.25]$, (b) ColorJintter: all the parameters are $0.15$ and (c) Horizontal Flip. For $\emph{V\textsubscript{MBFN}}$ in IITD, (a) RandomResizedCrop: scale and ratio are both $[0.9, 1.1]$, (b) ColorJintter: all the parameters are $0.1$, (c) RandomAffine: degress is $[-5, 5]$, translate is $[-0.05, 0.05]$, scale is $[0.9, 1,1]$ and shear is $0.1$ and (d) Horizontal Flip.}
\item{For $\emph{V\textsubscript{IR}}$ in PolyU, (a) RandomResizedCrop: scale and ratio are both $[0.9, 1.1]$, (b) ColorJintter: brightness is $0.3$, contrast is $0.25$, saturation and hue are both $0.15$ and (c) Horizontal Flip. For $\emph{V\textsubscript{MBFN}}$ in PolyU, (a) RandomResizedCrop: scale and ratio are both $[0.9, 1.1]$, (b) ColorJintter: brightness is $0.25$, the rest of parameters are all $0.15$ and (c) Horizontal Flip.}
\item{For $\emph{V\textsubscript{IR}}$ in TCD, (a) RandomResizedCrop: scale and ratio are both $[0.8, 1.2]$, (b) ColorJintter: brightness and hue are both $0.1$, the rest of parameters are $0.15$, (c) Horizontal Flip and (d) Gaussian Noise: mean is $7$, var is $0.1$. For $\emph{V\textsubscript{MBFN}}$ in TCD, (a) RandomResizedCrop: scale and ratio are both $[0.9, 1.1]$, (b) ColorJintter: all the parameters are $0.1$, (c) Horizontal Flip and (d) Gaussian Noise: mean is $7$, var is $0.1$.}
\item{For $\emph{V\textsubscript{IR}}$ in MPD, (a) RandomResizedCrop: scale and ratio are both $[0.75, 1.25]$, (b) ColorJintter: brightness is $0.25$, the rest of parameters are all $0.15$ and (c) Horizontal Flip. For $\emph{V\textsubscript{MBFN}}$ in MPD, (a) RandomResizedCrop: scale and ratio are both $[0.75, 1.25]$, (b) ColorJintter: brightness is $0.25$, the rest of parameters are all $0.1$ and (c) Horizontal Flip.}
\end{enumerate}
Ultimately, palmprint ROI images generated from 160 subjects were selected as the training set, and ROI images generated from the remaining 40 subjects formed the test set.

\section{Experimental Results}\label{sect:Experiments}
\subsection{Implementation Details of Detector $D\textsubscript{X}$s in DeepMPV+}\label{subsect:detectorDetails}
Five state-of-the-art or representative DCNN architectures, including PeleeNet+SSD \cite{wang2018pelee,liu2016ssd}, ShuffleNet+SSD \cite{ZhangArXiv2017}, MobileNetV1+SSD \cite{howard2017mobilenets}, MobileNetV2+SSD \cite{sandler2018inverted} and Tiny-YOLOV3\cite{yolov3}, were investigated in the phase of detecting double-finger-gap and palm-center, and the corresponding concrete models for detector $\emph{D\textsubscript{X}}$ are referred to as $\emph{D\textsubscript{PN}}$, $\emph{D\textsubscript{SN}}$, $\emph{D\textsubscript{MN1}}$, $\emph{D\textsubscript{MN2}}$ and $\emph{D\textsubscript{YOLO}}$, respectively.

\begin{table}[H]
 \caption{Settings for the key hyper-parameters used when training $\emph{D\textsubscript{X}}$s}
  \centering
  \begin{tabular}{cp{1.5cm}<{\centering}cp{1cm}<{\centering}p{1.5cm}<{\centering}}
    \toprule
    $\emph{D\textsubscript{X}}$s & Learning Rate & Optimizer & Batch Size & Weight Decay\\
    \midrule
    $\emph{D\textsubscript{PN}}$ & 0.005 & SGD & 32 & 0.0005\\
    $\emph{D\textsubscript{SN}}$ & 0.005 & RMSProp & 8 & 0.00005\\
    $\emph{D\textsubscript{MN1}}$ & 0.005 & RMSProp & 16 & 0.00005\\
    $\emph{D\textsubscript{MN2}}$ & 0.0005 & RMSProp & 8 & 0.0001\\
    $D\emph{\textsubscript{YOLO}}$ & 0.001 & - & 16 & 0.0005\\
    \bottomrule
  \end{tabular}
  \label{tab:settingsds}
\end{table}

$\emph{D\textsubscript{X}}$s were trained on $\rm{MPD\textsubscript{D}}$. For training $\emph{D\textsubscript{X}}$s, we used the fine-tuning strategy, i.g., $\emph{D\textsubscript{X}}$s were fine-tuned from the deep models pre-trained on MS-COCO or VOC0712 for the task of object detection. Except for YOLO, Caffe was used as our deep learning platform when training models. Settings for key hyper-parameters used when training $\emph{D\textsubscript{X}}$s, including the learning rate, the optimizer, the batch size and the weight decay, are summarized in Table \ref{tab:settingsds}. In Table \ref{tab:settingsds}, ``RMSProp'' and ``SGD'' are two different optimization methods implemented in Caffe.

\subsection{Implementation Details of Verifier $V\textsubscript{X}$s in DeepMPV+}\label{subsect:verifierDetails}
Taking ArcFace \cite{deng2019arcface} as the training framework, two representative DCNN architectures, IR\_50 \cite{szegedy2017inception} and MobileFaceNet \cite{chen2018mobilefacenets}, were selected as backbones, and the corresponding $\emph{V\textsubscript{X}}$ concrete models were called $\emph{V\textsubscript{IR}}$ and $\emph{V\textsubscript{MBFN}}$, respectively.

\begin{table}[H]
 \caption{Settings for the key hyper-parameters used when training $\emph{V\textsubscript{X}}$s}
  \centering
  \begin{tabular}{cp{1.5cm}<{\centering}p{1.5cm}<{\centering}cp{1.5cm}<{\centering}}
    \toprule
    $\emph{V\textsubscript{X}}$s & Learning Rate & Weight Decay & Epoch & Input Dimension \\
    \midrule
    $\emph{V\textsubscript{IR}}$ & 0.005 & 0.0005 & 26 & $224\times224$ \\
    $\emph{V\textsubscript{MBFN}}$ & 0.005 & 0.0005 & 26 & $224\times224$ \\
    \bottomrule
  \end{tabular}
  \label{tab:settingsvs}
\end{table}

$\emph{V\textsubscript{X}}$s were trained on multiple datasets including $\rm{MPD\textsubscript{ROI}}$. Pytorch was used as our deep learning platform when training models. Settings for key hyper-parameters used when training $\emph{V\textsubscript{X}}$s, including the learning rate, weight decay, epochs and input dimension, are summarized in Table \ref{tab:settingsvs}.

\subsection{Evaluation of Detector $D\textsubscript{X}$s on MPD}\label{subsect:evalDxs}

\begin{figure}
\centering
\includegraphics[width=1.0\columnwidth]{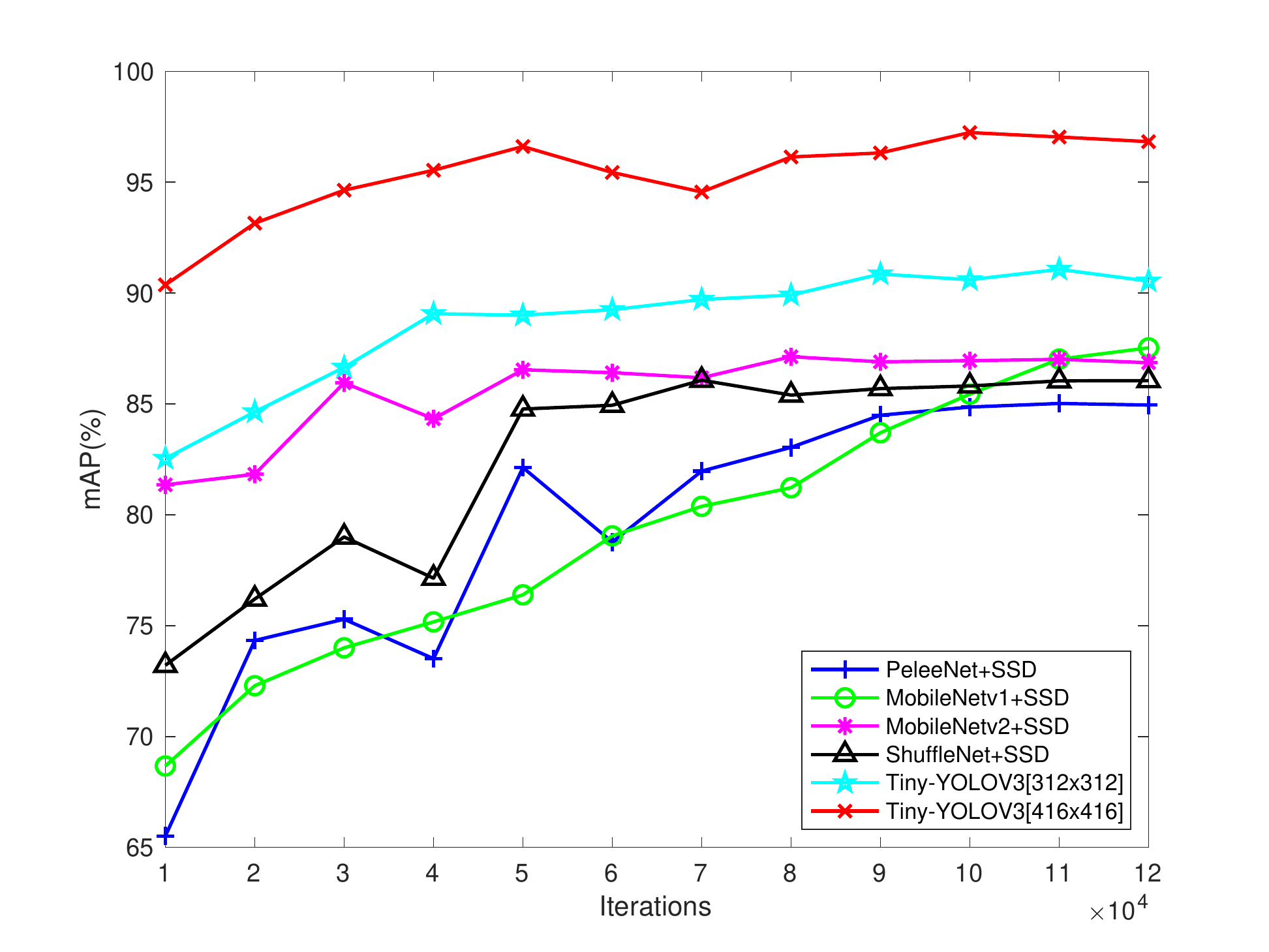}
\caption{The mAP change curve during the training of $\emph{D\textsubscript{X}}$s.}
\label{fig:5}
\end{figure}

\begin{table}[H]
  \caption{Sample table title}
   \centering
   \begin{tabular}{ccc}
   \toprule
   $\emph{D\textsubscript{X}}$s & Input Dimension & Time Cost (ms)\\
   \midrule
   $\emph{D\textsubscript{PN}}$ & 304$\times$304 & 96.7 \\
   $\emph{D\textsubscript{SN}}$ & 300$\times$300 & 85.8 \\
   $\emph{D\textsubscript{MN1}}$ & 300$\times$300 & 99.6 \\
   $\emph{D\textsubscript{MN2}}$ & 300$\times$300 & 105.5 \\
   $\emph{D\textsubscript{YOLO}}$ & 312$\times$312 & 91.9\\
   $\textbf{\emph{D\textsubscript{YOLO}}}$ & \textbf{416$\times$416} & \textbf{94.5}\\
   \bottomrule
 \end{tabular}
 \label{tab:resultsofcontrast}
\end{table}

In our palmprint verification system DeepMPV+, keypoint detection is a crucial step. As mentioned in Sect. \ref{subsect:detectorDetails}, the performance of five DCNN-based methods was evaluated. The change curves of the mAP during the training stage of 5 models are illustrated in Fig. \ref{fig:5}. As we can see, the mAP of $\emph{D\textsubscript{YOLO}}$ reaches 96\%, which is much higher than that of the other detectors. It is obvious that the Tiny-YOLOV3-based model converged much faster than the other models according to Fig. \ref{fig:5}.

All five models are implemented on iPhone 8 Plus, and the speed at which each image is processed is recorded.
Table \ref{tab:resultsofcontrast} lists the time costs of processing a single image.
To eliminate the influence of the input image size, all the images were resized to approximately 300$\times$300.
Although $\emph{D\textsubscript{YOLO}}$ takes 10 ms longer than $\emph{D\textsubscript{SN}}$ to detect a single image, it is still reasonable to select the former because its mAP value is nearly 10\% higher.
Comparing two Tiny-YOLOV3-based models with different input dimensions, the model with the input dimension $416\times416$ is slower but much more accurate.

\begin{figure}
  \centering
  \subfigure[]{
   \label{fig:6a}
   \includegraphics[width=2.5in]{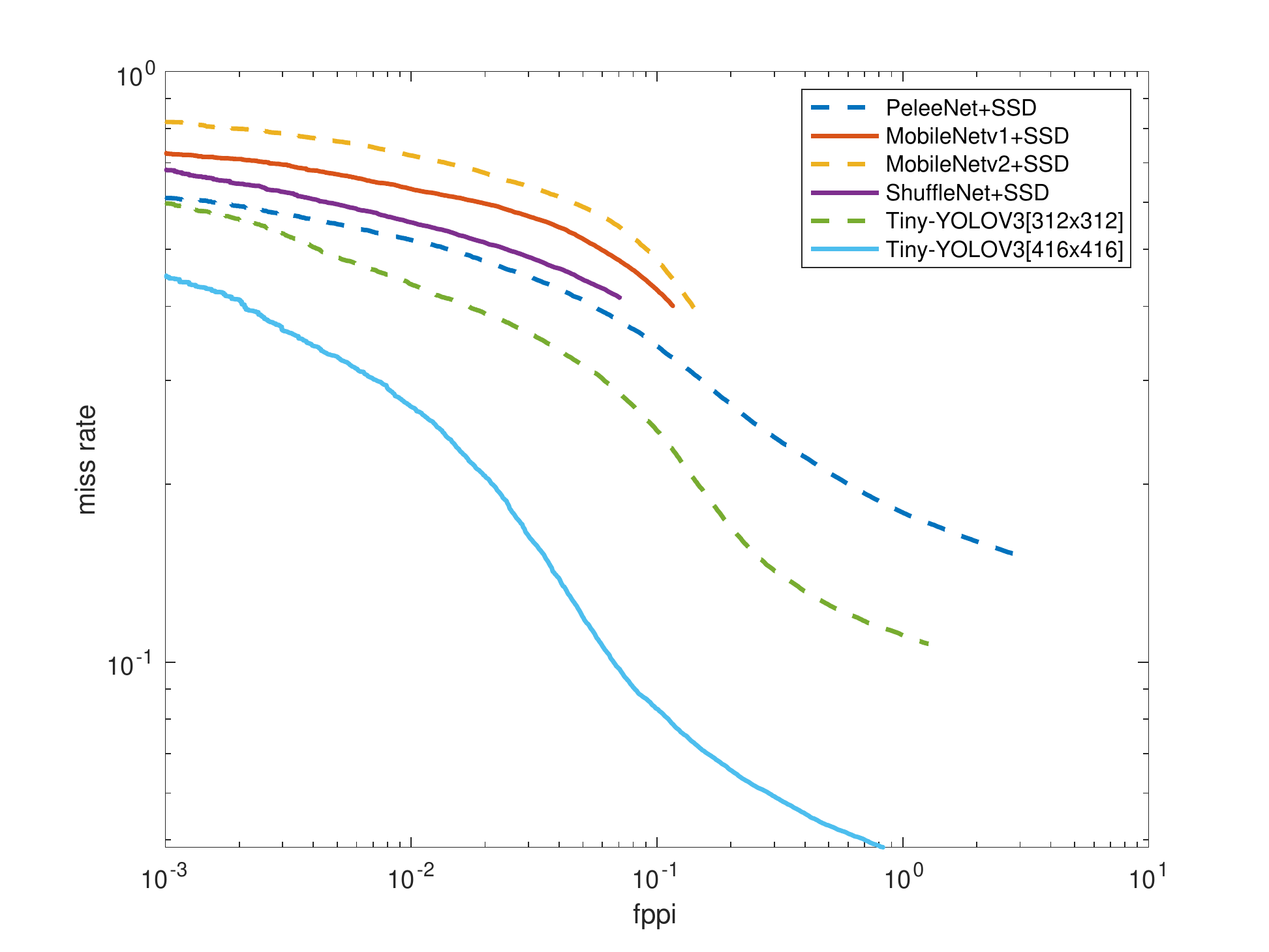}
   }
   \subfigure[]{
   \label{fig:6b}
   \includegraphics[width=2.5in]{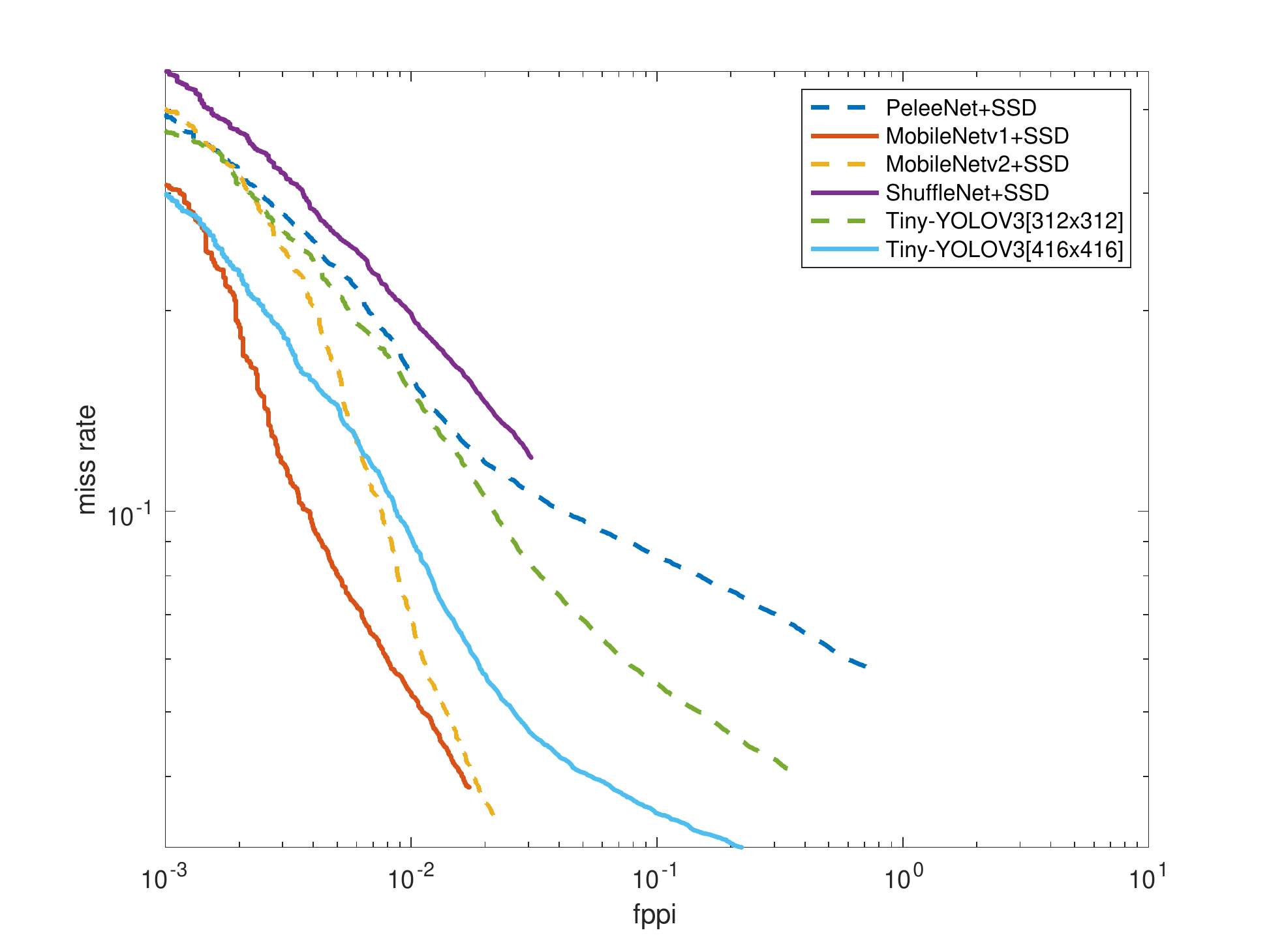}
   }
\caption{Keypoint detection results by different models. (a) shows the results of ``double-finger-gap'' while (b) shows the results of ``palm-center''.}
\label{fig:6}
\end{figure}

To better illustrate the performance of these 5 models, miss rate against false positives per image (FPPI) using log-log plots was plotted by varying the threshold on detection confidence. For a ground-truth keypoint $g\textsubscript{i}$ , if there is a detected keypoint $d\textsubscript{i}$ satisfying $\left \| {g\textsubscript{i} - d\textsubscript{i}} \right \| < \delta$ , where $\delta$ is a predefined threshold, we deemed that $g\textsubscript{i}$ is correctly detected and $d\textsubscript{i}$ is a true positive. In this experiment, $\delta$ was set as 10. The figures were plotted for the ``double-finger-gap'' (Fig. \ref{fig:6a}) and the ``palm-center'' (Fig. \ref{fig:6b}). As recommended in \cite{dollar2011pedestrian}, the log-average miss rate (LAMR) was used to summarize detector performance, computed by averaging miss rate at nine FPPI rates evenly spaced in log-space in the range $10^{-3}$ to $10^{1}$. The LAMRs achieved by different models are also shown in Fig. \ref{fig:6}.
It is obvious that the Tiny-YOLOV3-based model with the input dimension $416\times416$ has the best performance on double-finger-gap detection according to Fig. \ref{fig:6a}. As seen from Fig. \ref{fig:6b}, MobileNetV1 and MobileNetV2 have better performance on palm-center detection. However, overall, the models based on Tiny-YOLOV3 have excellent performance on both classes. In short, Tiny-YOLOV3 outperforms other models in terms of the overall performance on MPD.

\begin{table}
\center
\caption{5-fold cross validation of detector $\emph{D\textsubscript{YOLO}}$}
\begin{tabular}{cccc}
\hline
\hline
No. & AP of class 0 (\%) & AP of class 1 (\%) & mAP (\%)\\
\hline
1 & 99.76 & 100.00 & 99.88 \\
2 & 99.56 & 100.00 & 99.78 \\
3 & 99.66 & 100.00 & 99.83 \\
4 & 99.70 & 100.00 & 99.85 \\
5 & 96.69 & 98.10 & 97.39\\
\hline
Average & 99.05 & 99.62 & 99.35\\
\hline
\hline
\end{tabular}
\label{tab:crossvalidation}
\end{table}To better illustrate the performance of these 5 models, miss rate against false positives per image (FPPI) using log-log plots was plotted by varying the threshold on detection confidence. For a ground-truth keypoint $g\textsubscript{i}$ , if there is a detected keypoint $d\textsubscript{i}$ satisfying $\left \| {g\textsubscript{i} - d\textsubscript{i}} \right \| < \delta$ , where $\delta$ is a predefined threshold, we deemed that $g\textsubscript{i}$ is correctly detected and $d\textsubscript{i}$ is a true positive. In this experiment, $\delta$ was set as 10. The figures were plotted for the ``double-finger-gap'' (Fig. \ref{fig:6a}) and the ``palm-center'' (Fig. \ref{fig:6b}). As recommended in \cite{dollar2011pedestrian}, the log-average miss rate (LAMR) was used to summarize detector performance, computed by averaging miss rate at nine FPPI rates evenly spaced in log-space in the range $10^{-3}$ to $10^{1}$. The LAMRs achieved by different models are also shown in Fig. \ref{fig:6}.
It is obvious that the Tiny-YOLOV3-based model with the input dimension $416\times416$ has the best performance on double-finger-gap detection according to Fig. \ref{fig:6a}. As seen from Fig. \ref{fig:6b}, MobileNetV1 and MobileNetV2 have better performance on palm-center detection. However, overall, the models based on Tiny-YOLOV3 have excellent performance on both classes. In short, Tiny-YOLOV3 outperforms other models in terms of the overall performance on MPD.

\begin{table}[H]
\caption{5-fold cross validation of detector $\emph{D\textsubscript{YOLO}}$}
\centering
\begin{tabular}{cccc}
\toprule
No. & AP of class 0 (\%) & AP of class 1 (\%) & mAP (\%)\\
\midrule
1 & 99.76 & 100.00 & 99.88 \\
2 & 99.56 & 100.00 & 99.78 \\
3 & 99.66 & 100.00 & 99.83 \\
4 & 99.70 & 100.00 & 99.85 \\
5 & 96.69 & 98.10 & 97.39\\
\midrule
Average & 99.05 & 99.62 & 99.35\\
\bottomrule
\end{tabular}
\label{tab:crossvalidation}
\end{table}

To obtain a reliable and stable model, $\emph{D\textsubscript{YOLO}}$ was cross-validated using 5-fold cross validation. Class 0 and class 1 in Table \ref{tab:crossvalidation} indicate the class ``double-finger-gap'' and the class ``palm-center'', respectively. According to Table \ref{tab:crossvalidation}, $\emph{D\textsubscript{YOLO}}$ performs well on our proposed MPD dataset.

In conclusion, we choose the Tiny-YOLOV3-based model with an input dimension of $416\times416$ as our final detector $\textbf{\emph{D}}$.

\subsection{Evaluation of $V\textsubscript{X}$s on Multiple Datasets}\label{subsect:performMbfn}

\begin{table*}
  \caption{Publicly Available Datasets Used}
  \centering
  \begin{tabular}{p{2.6cm}p{0.5cm}<{\centering}p{0.5cm}p{1.5cm}p{7cm}p{1cm}<{\centering}p{0.8cm}<{\centering}p{1cm}<{\centering}}
    \toprule
    Name & Ref. & Year & Capture type & Features & N. img. & N. pal. & N. sess. \\
    \midrule
    CASIA & \cite{CASIA} & 2007 & Contactless & Black background, uniform illumination conditions, slight hand pose variation and multi-spectrum. & 5,502 & 624 & 1\\
    IIT-D v1 & \cite{IIT} & 2007 & Contactless & Uniform background, uniform illumination conditions, slight hand pose variation. & 2,601 & 460 & 1\\
    PolyU \uppercase\expandafter{\romannumeral2} Palmprint Dataset & \cite{PolyU} & 2011 & Contactless & Uniform background, uniform illumination conditions, 5 hand postures, right hand, both 3D and 2D images. & 1,140 & 114 & 1\\
    Tongji Contactless Palmprint Dataset & \cite{TongjiPD} & 2017 & Contactless & Uniform background, uniform illumination conditions, constrained hand posture. & 12,000 & 600 & 2\\
    MPD & - & 2020 & Smartphones & Uncontrolled background, multiple illumination conditions, various hand postures. & 16,000 & 400 & 2\\
    \bottomrule
  \end{tabular}
  \label{tab:publicdatasets}
Notes: N. img. = Number of images; N. pal. = Number of palms; N. sess. = Number of sessions.
\end{table*}

To evaluate the performance of our proposed method, experiments were carried out on 5 contactless public palmprint image datasets including MPD. All these datasets are listed in detail in Table \ref{tab:publicdatasets}. For the convenience of recording, the PolyU \uppercase\expandafter{\romannumeral2} Palmprint Dataset will be recorded as PolyU and the Tongji Contactless Palmprint Dataset as TCD. For every dataset, the left and right palms of the same person are considered as belonging to different individuals.
In addition, in the process of study, we considered using SiameseMobileNet \cite{zhang2019pay} to match palmprints, which was eventually replaced by ArcFace$+$MobileFaceNet with better performance. More importantly, we compared the performance of the proposed method and the most recent methods in the literature on these datasets.
Since most of the methods in the literature do not publish their source code, we re-implemented algorithms in some papers.
To ensure the fairness of the comparison, we only compared methods with the same evaluation procedure. In addition, all experimental data of $\emph{V\textsubscript{X}}$s are the average values after 5-fold cross validation to avoid bias.

\subsubsection{Recognition Accuracy}
At this stage, we chose to use the Top-1 Accuracy to evaluate the performance of all these methods. Specifically, for each palm (each ID), we randomly select an image as the registration image, take the rest of the images as input, and compare the selected image with the registration images one by one to find the ID of the input image. The Top-1 Accuracy is highly consistent with the real application scene, and it can represent the recognition performance of the methods well.

The recognition accuracy of $\emph{V\textsubscript{X}}$s is compared with that of the most recent methods reported in the literature. To compare methods based on local texture descriptors, CompCode \cite{kong2004competitive}, OLOF \cite{sun2005ordinal}, CR-CompCode \cite{zhang2017towards}, DDBC \cite{fei2019learning} and ALDC\_M \cite{fei2019local} are considered. To compare methods based on deep learning, we considered FERNet \cite{matkowski2019palmprint}, PalmNet \cite{genovese2019palmnet},
VGG-16 \cite{tarawneh2018pilot}, AlexNet-S \cite{ramachandra2018verifying}, ResNet-50 \cite{he2016deep}, GoogLeNet \cite{szegedy2015going} and PCANet \cite{chan2015pcanet}. Table \ref{tab:resultsofTop1} describes the Top-1 accuracy of $\emph{V\textsubscript{X}}$s and the methods in the literature. As we can see, $\emph{V\textsubscript{IR}}$'s on the considered public palmprint datasets outperforms all these methods in the literature. In particular, $\emph{V\textsubscript{IR}}$'s Top-1 accuracy on the PolyU \uppercase\expandafter{\romannumeral2} reached 100\%. Additionally, the Top-1 accuracy of $\emph{V\textsubscript{MBFN}}$ in each dataset exceeded or approached that of the methods in the literature.

\begin{table}[H]
  \caption{Top-1 Accuracy (\%) of $\emph{V\textsubscript{X}}$s and other methods in literature}
  \begin{tabular}{p{1.8cm}ccccc}
    \toprule
    Method & CASIA & IIT-D & PolyU \uppercase\expandafter{\romannumeral2} & TCD & MPD\\
    \midrule
    PalmNet-GaborPCA* & 97.17 & 97.31 & 99.95 & 99.89 & 91.88 \\
    FERNet & 97.65 & 99.61 & 99.77 & 98.63 & -\\
    VGG-16 & 97.80 & 93.64 & - & 98.46 & -\\
    AlexNet-S & 92.76 & 97.24 & 70.96 & - & -\\
    ResNet-50 & 95.21 & 95.57 & 54.92 & - & -\\
    GoogLeNet & 93.84 & 96.22 & 68.59 & - & -\\
    PCANet & 95.53 & 97.37 & 99.66 & - & -\\
    DDBC & 96.41 & 96.44 & - & 98.73 & -\\
    ALDC\_M & 94.64 & 97.00 & 99.85 & - & -\\
    CompCode & 79.27 & 77.79 & 99.21 & - & -\\
    OLOF & 73.32 & 73.26 & 99.55 & - & -\\
    CR-CompCode & 91.73 & 94.22 & 97.18 & - & -\\
    Siamese- MobileNet & 53.47 & 59.30 & 86.76 & 92.98 & 78.53 \\
    $\textbf{\emph{V\textsubscript{IR}}}$  & $\textbf{98.91}$ & $\textbf{99.85}$ & $\textbf{100.00}$ & $\textbf{99.90}$ & $\textbf{99.18}$ \\
    $\textbf{\emph{V\textsubscript{MBFN}}}$ & $\textbf{98.11}$ & 99.19 & $\textbf{99.97}$ & 99.81 & $\textbf{98.06}$ \\
    \bottomrule
  \end{tabular}
*Re-implemented using official public code.
\label{tab:resultsofTop1}
\end{table}

\subsubsection{Verification Accuracy}
In this phase, we chose to use the Equal Error Rate (EER) to evaluate the performance of these methods because it is one of the most commonly used metrics for biometric verification.
We compared $\emph{V\textsubscript{X}}$s with learning-based methods such as FERNet \cite{matkowski2019palmprint} and PalmNet \cite{genovese2019palmnet}, VGG-16 \cite{tarawneh2018pilot}, AlexNet-S \cite{ramachandra2018verifying}, ResNet-50 \cite{he2016deep}, GoogLeNet \cite{szegedy2015going} and PCANet \cite{chan2015pcanet}, and local-texture-descriptor-based methods like CompCode \cite{kong2004competitive}, OLOF \cite{sun2005ordinal} and CR-CompCode \cite{zhang2017towards}.
Table \ref{tab:resultsofEER} details the EERs of $\emph{V\textsubscript{X}}$s and the methods in the literature.
It can be seen from Table \ref{tab:resultsofEER} that the performance of $\emph{V\textsubscript{IR}}$ on these public palmprint datasets is better than that of all the methods in the literature, while the EER of $\emph{V\textsubscript{MBFN}}$ on each dataset exceeded or approached the methods in the literature.

\begin{table}
  \caption{EERs (\%) of $\emph{V\textsubscript{X}}$s and other methods in literature}
  \centering
  \begin{tabular}{p{1.8cm}ccccc}
    \toprule
    Method & CASIA & IIT-D & PolyU \uppercase\expandafter{\romannumeral2} & TCD & MPD\\
    \midrule
    PalmNet- GaborPCA* & 3.21 & 3.83 & 0.39 & 0.40 & 6.22\\
    FERNet  & 0.73 & 0.76 & 0.15 & - & -\\
    VGG-16 & 7.86 & 7.44 & - & 2.86 & -\\
    AlexNet-S & 1.79 & 0.92 & 14.91 & - & -\\
    ResNet-50 & 4.27 & 3.68 & 13.45 & - & -\\
    GoogLeNet & 1.65 & 1.97 & 11.19 & - & -\\
    PCANet & 1.46 & 1.18 & 0.45 & - & -\\
    CompCode & 1.08 & 1.39 & 0.68 & - & -\\
    OLOF & 1.75 & 2.09 & 0.23 & - & -\\
    CR-CompCode & 3.18 & 2.78 & 1.02 & - & -\\
    Siamese- MobileNet & 9.72 & 5.24 & 2.18 & 1.38 & 3.03\\
    $\textbf{\emph{V\textsubscript{IR}}}$ & $\textbf{0.59}$ & $\textbf{0.47}$ & $\textbf{0.08}$ & $\textbf{0.21}$ & $\textbf{0.81}$ \\
    $\textbf{\emph{V\textsubscript{MBFN}}}$ & 0.85 & $\textbf{0.74}$ & 0.19 & $\textbf{0.34}$ & $\textbf{1.28}$ \\
    \bottomrule
  \end{tabular}
*Re-implemented using official public code
\label{tab:resultsofEER}
\end{table}

\subsubsection{Time Cost}

\begin{table}
  \caption{The average feature extraction (per image) and comparison time (ms) of different methods}
  \centering
  \begin{tabular}{lcp{2cm}<{\centering}p{2cm}<{\centering}}
    \toprule
    Method & Ref. & Feature Extraction & Comparison \\
    \midrule
    PalmNet-GaborPCA & \cite{genovese2019palmnet} & 1600.0000 & 10.0000 \\
    DDBC & \cite{fei2019learning} & 26.5000 & 2.0000 \\
    ALDC & \cite{fei2019local} & 78.5000 & 0.1000 \\
    $\emph{V\textsubscript{IR}}$ & - & 1230.6132 & $\textbf{0.0018}$ \\
    $\emph{V\textsubscript{MBFN}}$ & - & 100.4943 & $\textbf{0.0018}$ \\
    \bottomrule
  \end{tabular}
\label{tab:resultsofTime}
\end{table}

We measured the time consumed by each method to extract features from a single palmprint image and compare a pair of palmprint images, which are listed in Table \ref{tab:resultsofTime}. ArcFace calculates the similarity of a pair of palmprint features by a multiplicative way, which is very fast and takes approximately $1.77 \times 10^{-6}$s, almost negligible.
The speed of each method in Table \ref{tab:resultsofTime} in feature extraction can be roughly divided into two levels: 100ms level (DDBC, ALDC and $\emph{V\textsubscript{MBFN}}$) and 1000ms level (PalmNet and $\emph{V\textsubscript{IR}}$). Considering the memory limitation and speed requirement of smartphones, we chose the final verifier in 100ms level, and finally chose $\emph{V\textsubscript{MBFN}}$ with the best performance in 100ms level as our final verifier $\textbf{\emph{V}}$.

\subsubsection{Determination of Threshold}

\begin{table}
  \caption{TPRs (\%) with FAR=$10^{-1}$, FAR=$10^{-2}$, FAR=$10^{-3}$ and FAR=$10^{-4}$ of $\emph{V\textsubscript{X}}$s and other methods on MPD}
  \centering
  \begin{tabular}{p{1.5cm}p{1.3cm}<{\centering}p{1.3cm}<{\centering}p{1.3cm}<{\centering}p{1.3cm}<{\centering}}
    \toprule
    Method & FAR=$10^{-1}$ & FAR=$10^{-2}$ & FAR=$10^{-3}$ & FAR=$10^{-4}$ \\
    \midrule
    PalmNetGa- borPCA & 95.17 & 88.02 & 81.64 & 76.28\\
    SiameseMo- bileNet & 99.62 & 88.60 & 51.89 & 15.56\\
    $\textbf{\emph{V\textsubscript{IR}}}$ & $\textbf{99.82}$ & $\textbf{99.28}$ & $\textbf{96.53}$ & $\textbf{94.20}$\\
    $\textbf{\emph{V\textsubscript{MBFN}}}$ & $\textbf{99.66}$ & $\textbf{98.49}$ & $\textbf{94.97}$ & $\textbf{89.53}$\\
    \bottomrule
  \end{tabular}
\label{tab:resultsofFARTPR}
\end{table}

Finally, we used True Positive Rate (TPR)-False Acceptance Rate (FAR) to determine the matching threshold of DeepMPV+.
We tested TPR-FAR on MPD, which is the most complex dataset at present.
Table \ref{tab:resultsofFARTPR} details the TPR-FAR for $\emph{V\textsubscript{X}}$s and other methods.
It can be seen from Table \ref{tab:resultsofFARTPR} that $\emph{V\textsubscript{X}}$s are obviously superior to other methods. According to this table, for the sake of security, we chose the threshold corresponding to $\emph{V\textsubscript{MBFN}}$ when FAR=$10^{-4}$ as the final threshold $T$ to build a smartphone palmprint verification application, $T=0.5014$.

\subsection{Application System Implementation}

\begin{figure*}[t]
  \centering
  \subfigure[]{
   \label{fig:7a}
   \includegraphics[width=1.2in]{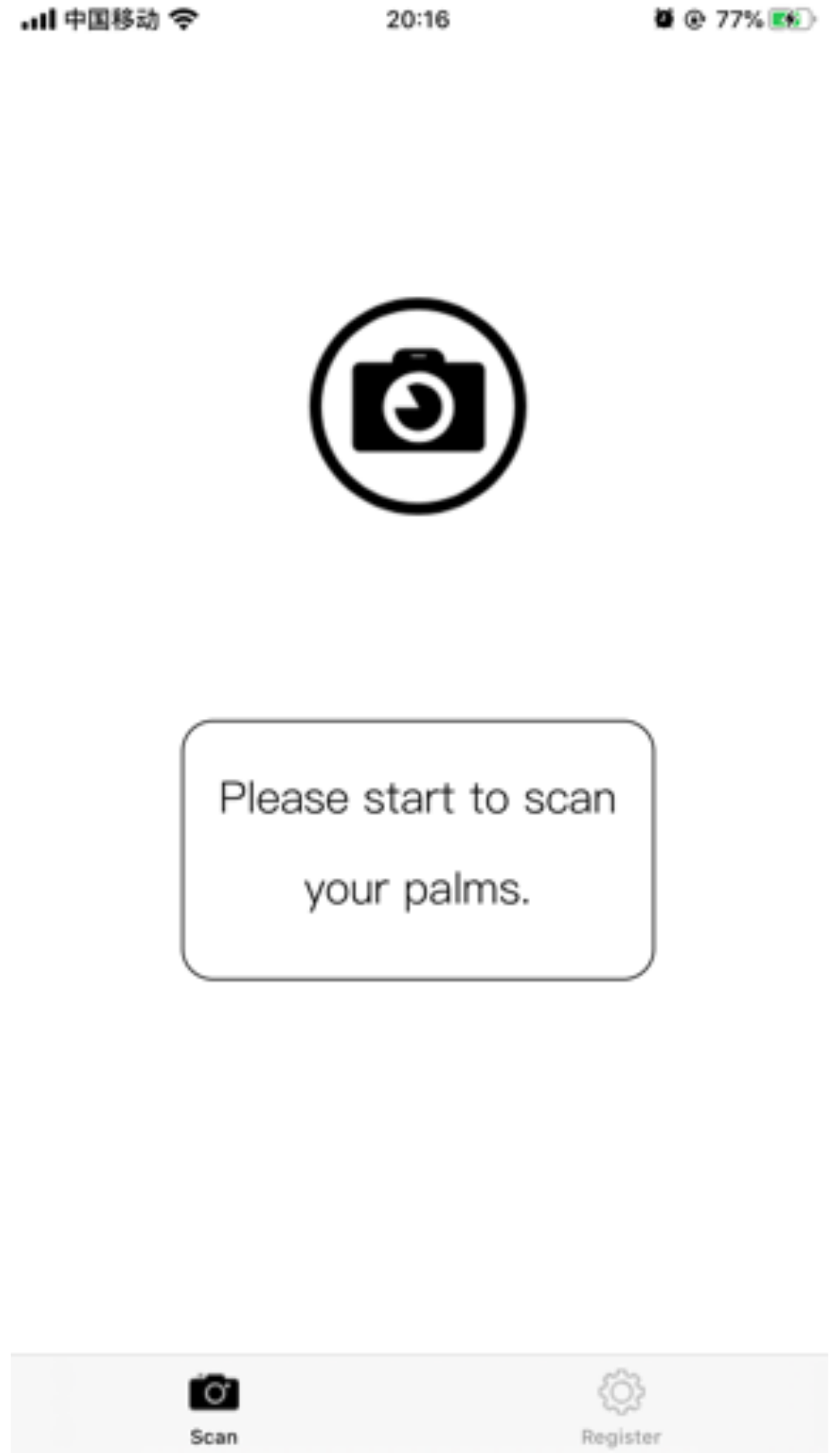}}
   \subfigure[]{
   \label{fig:7b}
   \includegraphics[width=1.2in]{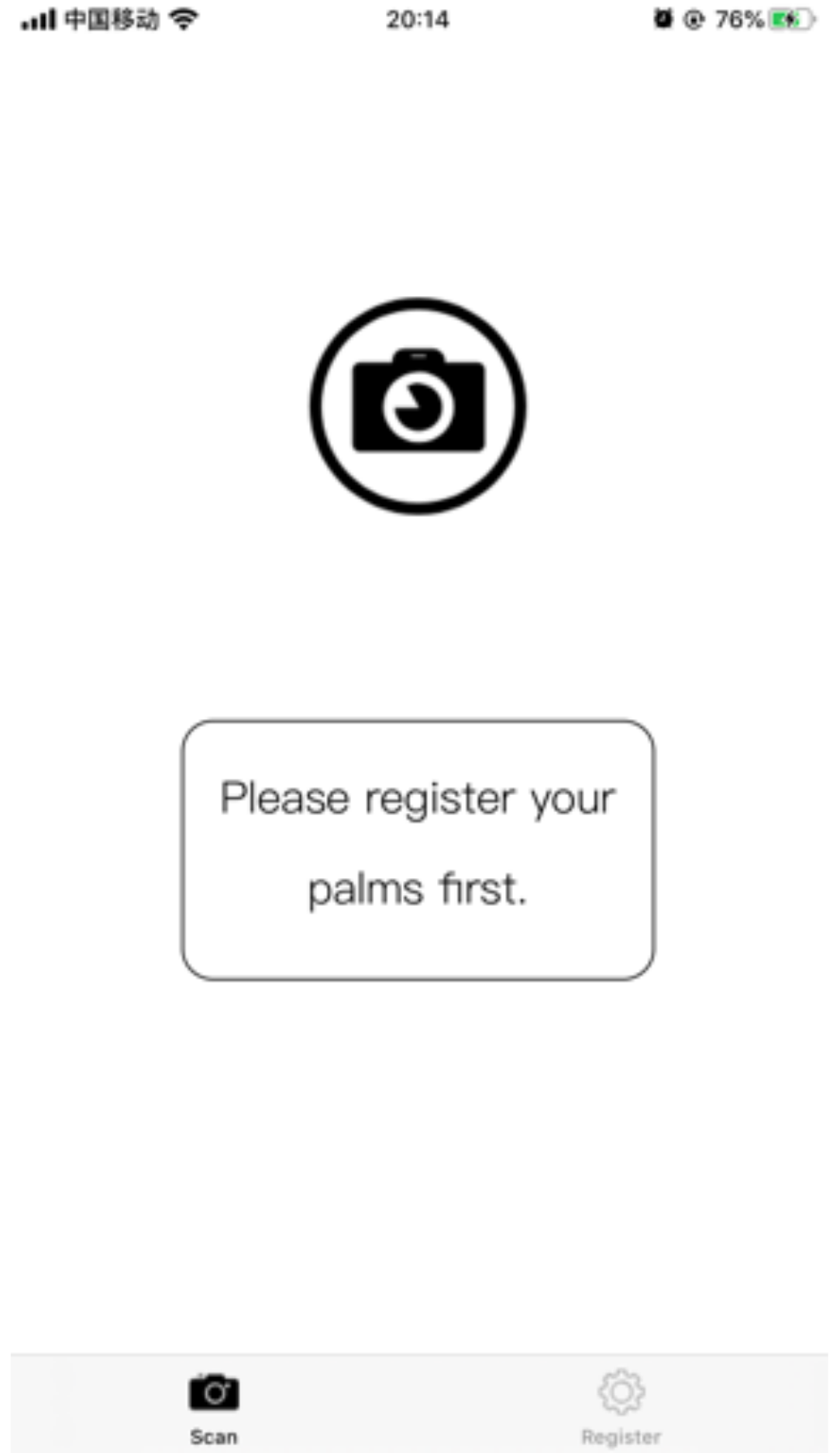}}
   \subfigure[]{
   \label{fig:7c}
   \includegraphics[width=1.2in]{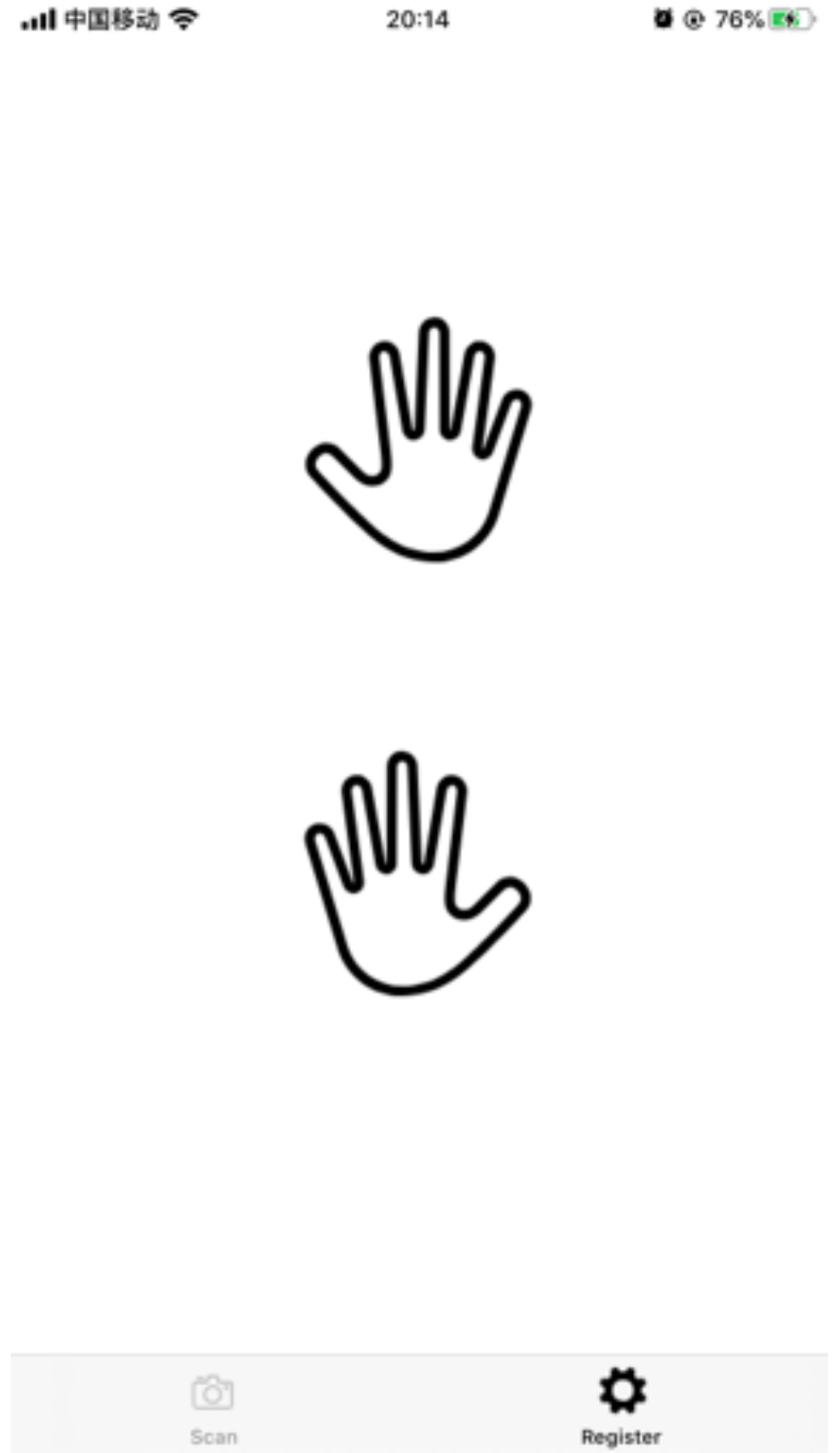}}
   \subfigure[]{
   \label{fig:7d}
   \includegraphics[width=1.2in]{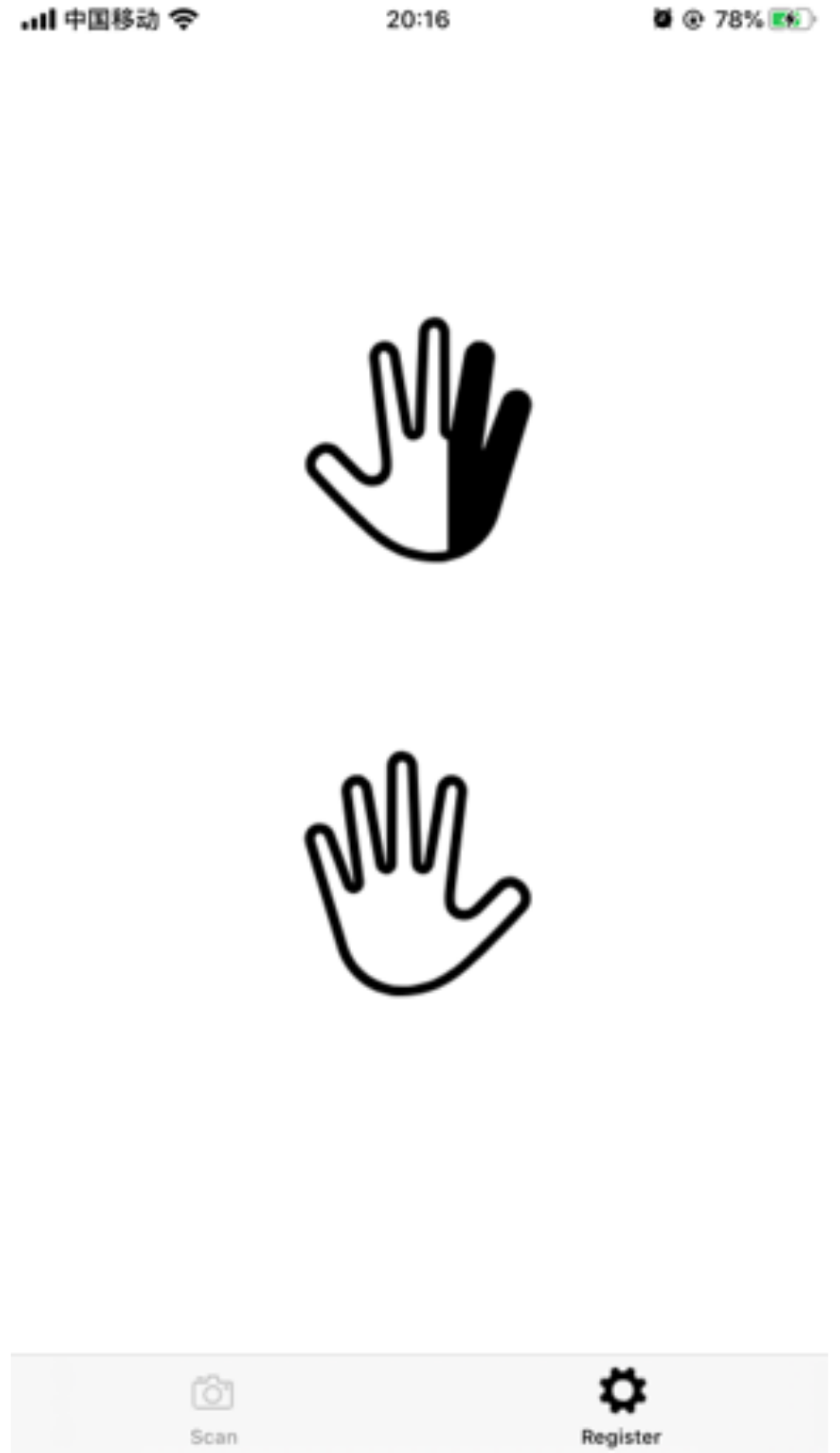}}
   \subfigure[]{
   \label{fig:7e}
   \includegraphics[width=1.2in]{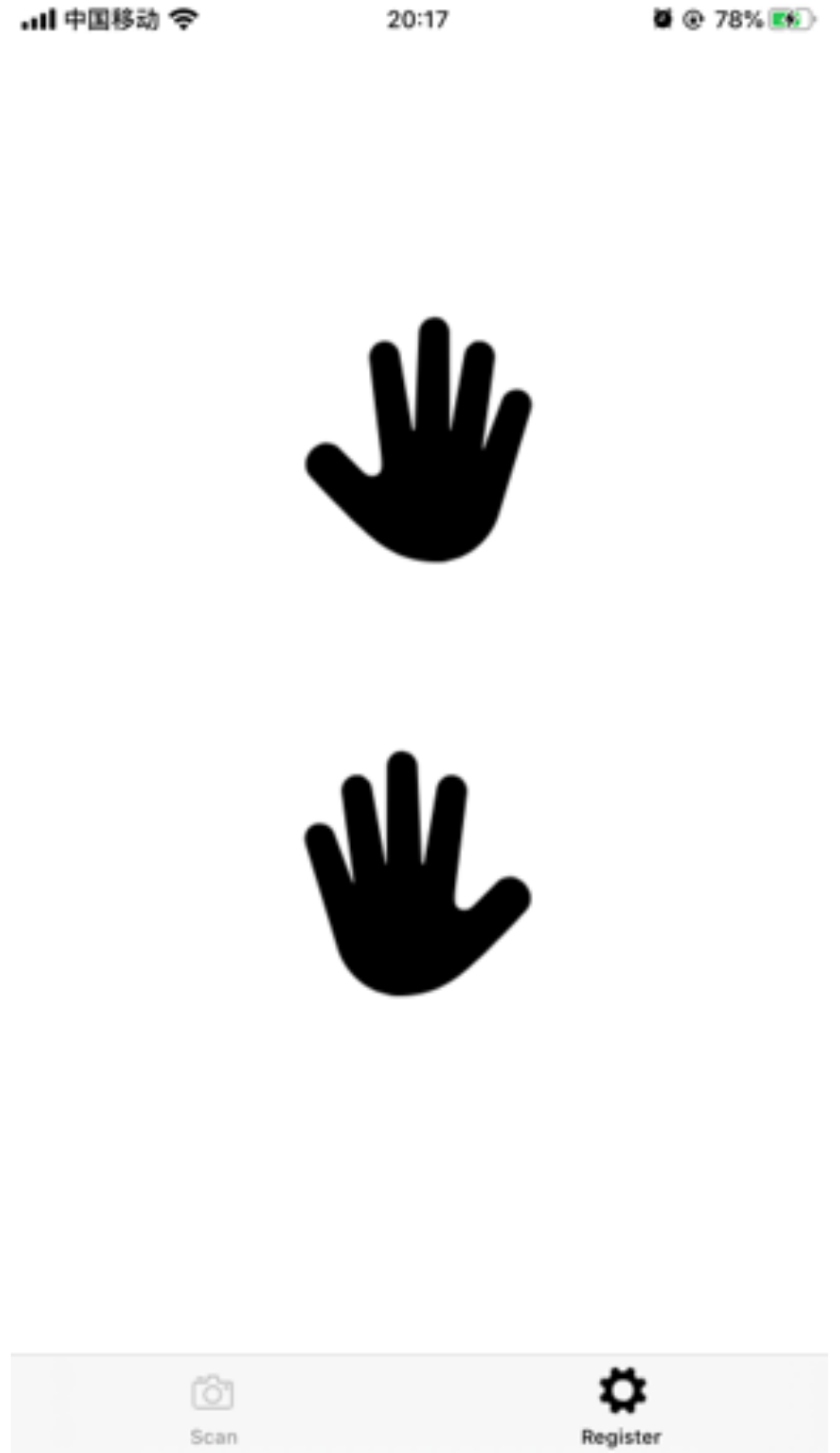}}
   \hspace{1in}
   \subfigure[]{
   \label{fig:7f}
   \includegraphics[width=1.2in]{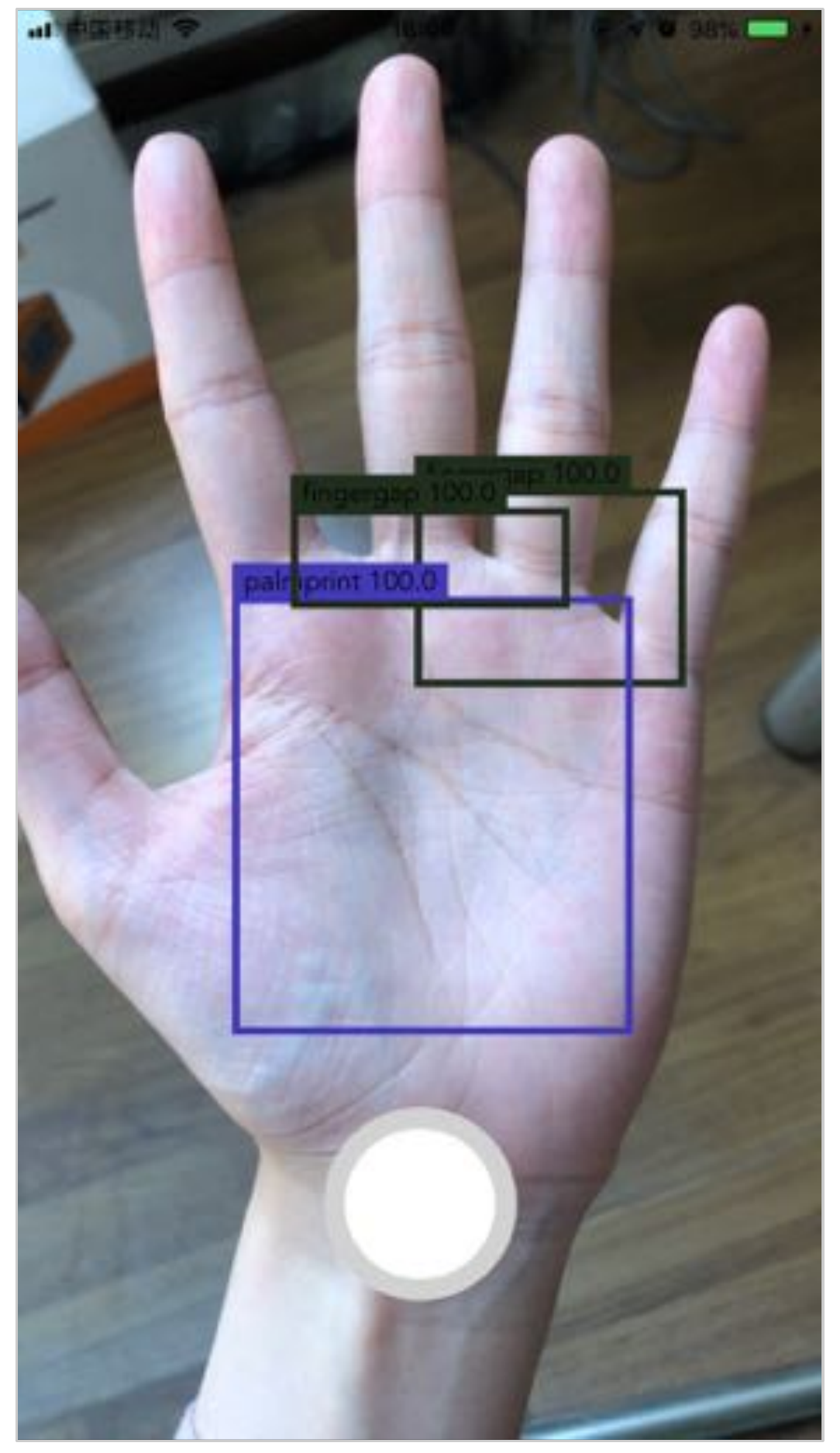}}
   \subfigure[]{
   \label{fig:7g}
   \includegraphics[width=1.2in]{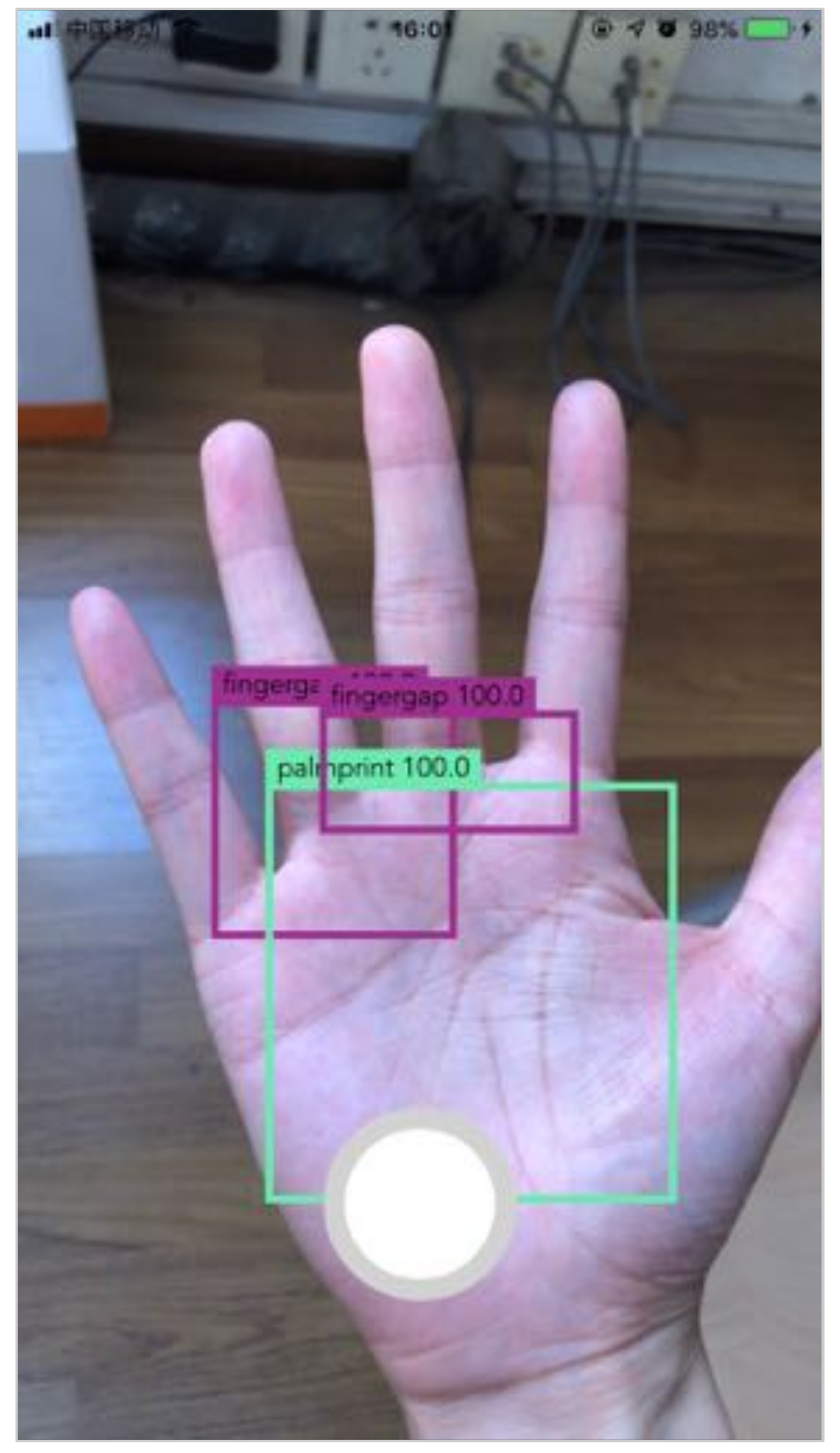}}
   \subfigure[]{
   \label{fig:7h}
   \includegraphics[width=1.2in]{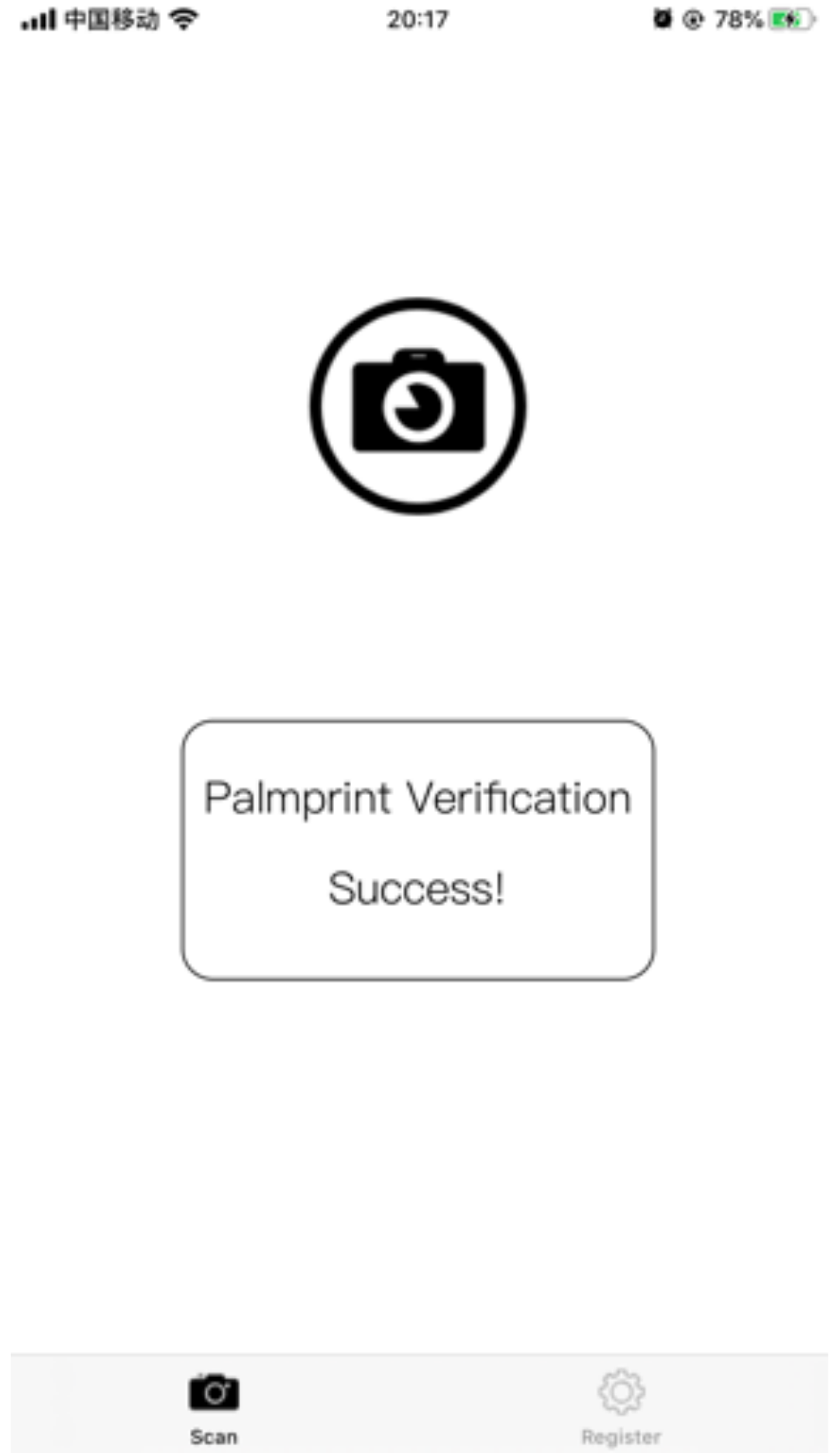}}
   \subfigure[]{
   \label{fig:7i}
   \includegraphics[width=1.2in]{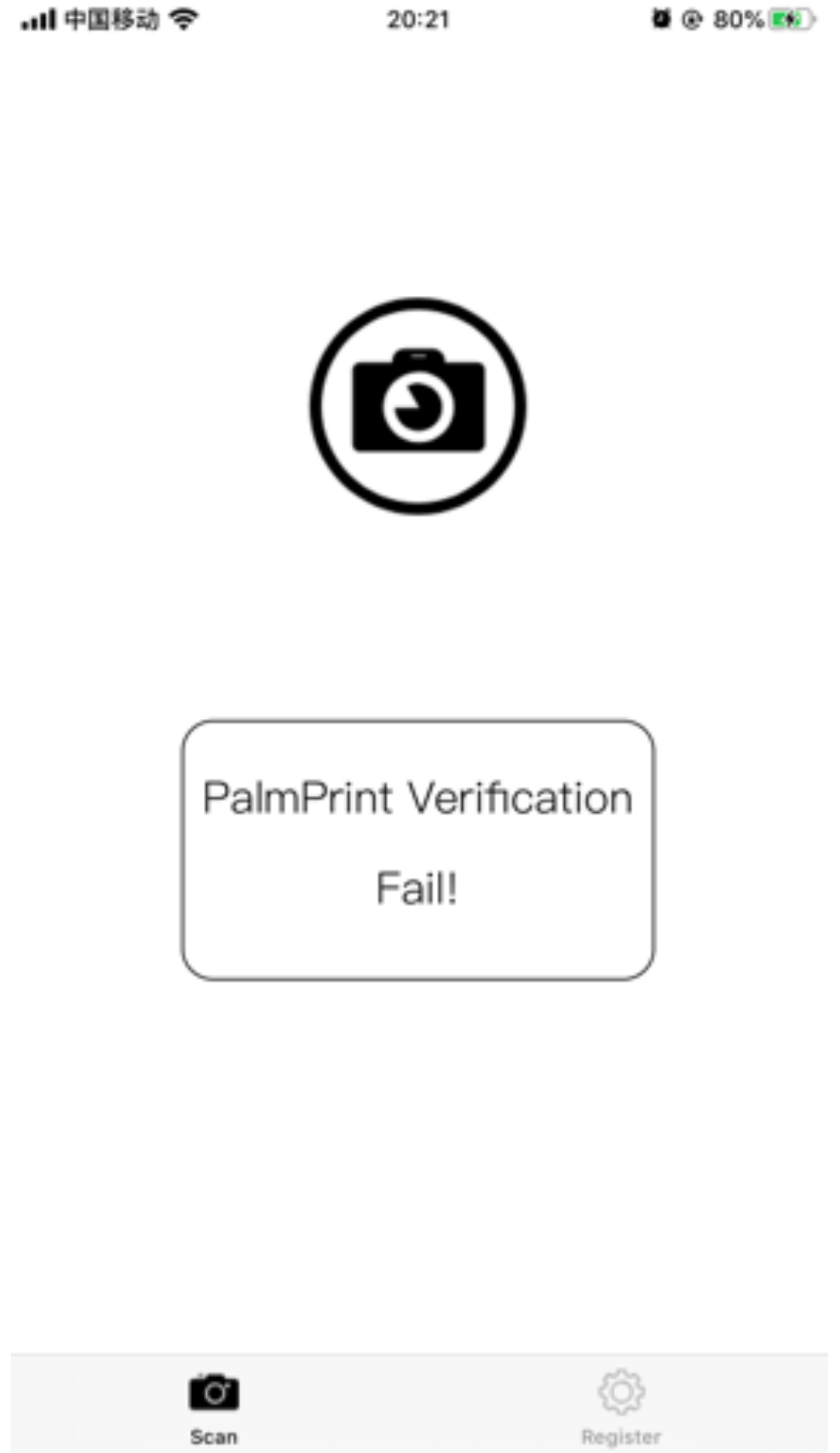}}
   \subfigure[]{
   \label{fig:7j}
   \includegraphics[width=1.2in]{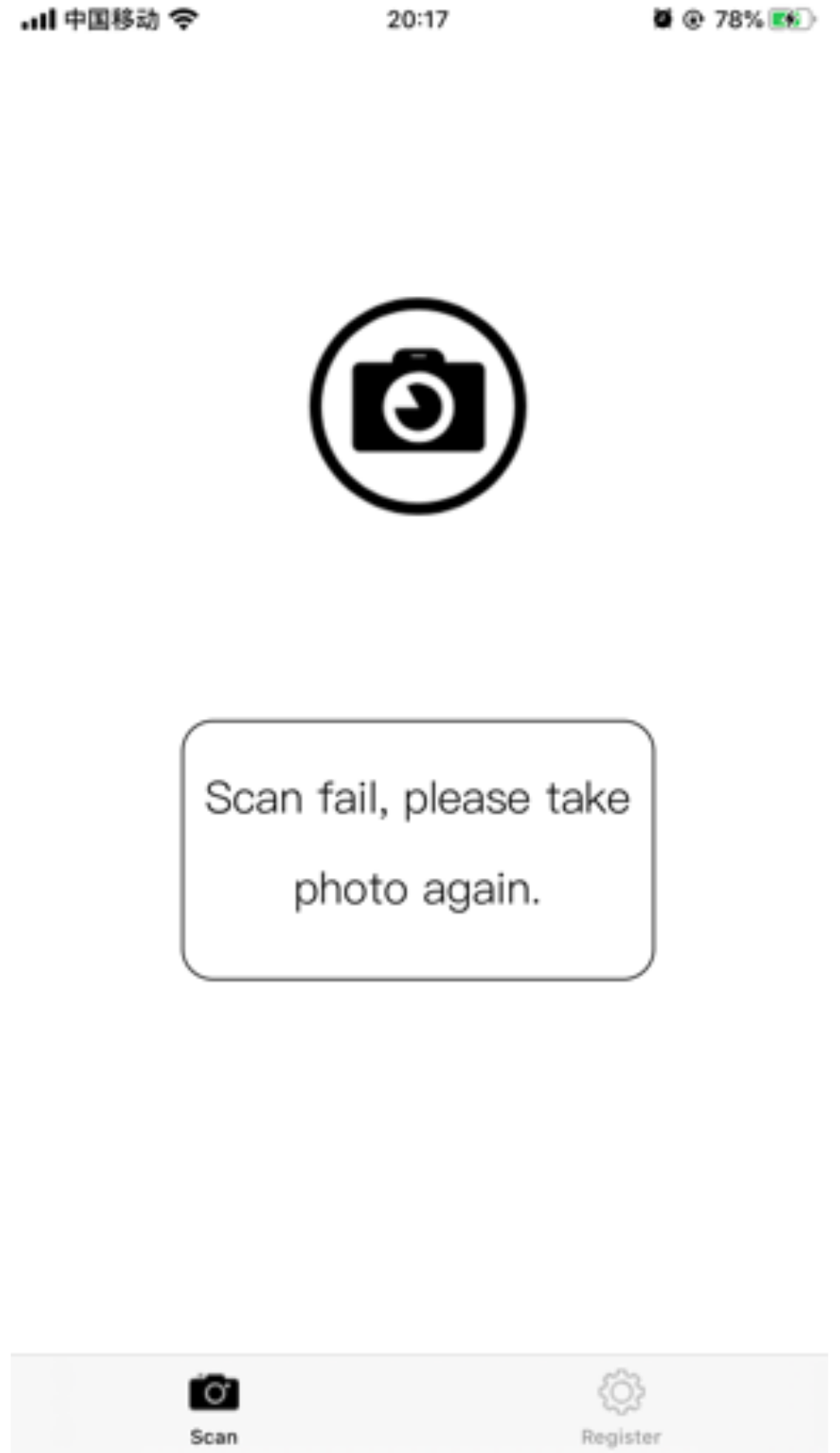}}
\caption{(a)$\sim$(j) are 10 screenshots of our palmprint verification application. (a) is the home and (b) is the hint page that suggests users to set up their palm images; (c), (d) and (e) are registration pages while (f) and (g) are the pages dynamically showing the process of ROI detection; (h), (i) and (j) are the pages showing the ultimate verification results.}
\label{fig:7}
\end{figure*}

A practical palmprint verification system was implemented as an iOS application on iPhone 8 Plus.
Illustrations of its interfaces are shown in Fig. \ref{fig:7}.
We will introduce our interfaces according to the user's steps.

When a user enters our application, the first page he or she will see is the home page, as shown in Fig. \ref{fig:7a}. The user then simply clicks the camera button in the center of the home page to scan his or her palmprint for authentication. However, if the user is using our application for the first time, it means he/she may not have registered his/her palmprint yet. We will remind the user to register his/her palmprint as shown in Fig. \ref{fig:7b}.

Before matching palmprints, the user's palmprint ROI images need to be stored in our local database. As we can see, Fig. \ref{fig:7c}$\sim$\ref{fig:7e} are the registration pages of our system. When the user enters the registration page for the first time, he/she will see the page with blank hand buttons as shown in Fig. \ref{fig:7c}. To avoid the impact of poorly photographed palm images, 3 palmprint ROI images are stored for each palm. When the user stores one palmprint ROI image, one-third of the palm buttons turn black, as shown in Fig. \ref{fig:7d}. Therefore, if the user has completed the registration of personal palmprint information, the hand button will turn to completely black, as illustrated in Fig. \ref{fig:7e}. If the user wants to reset his/her palm information, he/she only needs to click the hand button and re-store 3 palm images. The button will return to the blank state and change to the full black state during the user's reset operation.

When the user completes the registration of his/her palmprint information, he/she can use the camera to take palm images and match palmprints. If the user clicks the camera button, he/she will come to the scan pages like Fig. \ref{fig:7f} and Fig. \ref{fig:7g}. In general, our system will detect 2 double-finger-gaps and 1 palm-center in one hand. Fig. \ref{fig:7f} is a scan screenshot of the left hand while Fig. \ref{fig:7g} is a scan screenshot of the right hand. The user clicks the white circle button to save the photo of his/her palm. Real-time detection boxes can help users find better positions and angles for photo capture.

Once the user clicks the white circle button, our system will capture palm images, detect feature regions to construct a local coordinate system and extract palmprint ROIs.
The ROI image will be compared with pre-stored palmprint ROI images, and the highest score for palmprint verification will be selected. If the score is greater than $T$, the system will go back to the home page and give the feedback ``Palmprint Verification Success'' (as shown in Fig. \ref{fig:7h}); otherwise, the system will give the feedback ``Palmprint Verification Fail'' (as shown in Fig. \ref{fig:7i}).

In the phase of scanning palmprint, if the palmprint ROI cannot be extracted due to poor photography of the palmprint, our system will return to the home page and tell the user ``Scan fail, please take photo again''.

\section{Conclusion and Future Work}\label{sect:Conclusion}
In this paper, we made two major contributions to the field of vision-based palmprint verification.
First, we collected and labeled a large-scale palmprint dataset including 16,000 palm images from 200 subjects using multi-brand smartphones, which is the largest one in this field.
We have made it publicly available. Such a dataset will surely benefit the study of palmprint verification. Second, we proposed a DCNN-based solution for palmprint verification on mobile platforms.
Its high efficiency and efficacy have been corroborated by experiments.
Extensive experiments conducted on palmprint verification indicate that our proposed model DeepMPV+ can surpass all its competitors.

In the near future, we will try to refine our DCNN-based palmprint verification solution and to continuously enlarge MPD to include more palmprint samples. Furthermore, we will also try to improve our method by incorporating few-shot learning.





\end{document}